\documentclass{article}
\usepackage{graphicx}
\usepackage{booktabs}
\usepackage{adjustbox}
\usepackage{nicefrac}
\usepackage{indentfirst}
 \usepackage[table,xcdraw]{xcolor}
\usepackage{url}
\usepackage[a4paper, total={6in, 8in}]{geometry}
\usepackage{natbib}
\usepackage{color}
\usepackage{tabularray}
\usepackage{caption}
\usepackage{subcaption}
\usepackage{amsmath}
\usepackage{array}
\usepackage{amssymb,amsfonts}%
\usepackage{amsthm}%
\usepackage{enumitem}
\usepackage{multirow}
\usepackage{rotating}
\usepackage{bm}
\usepackage{setspace}

\definecolor{darkgreen}{rgb}{0,0.5,0}
\usepackage[colorlinks,allcolors=darkgreen]{hyperref}

\usepackage{nameref}

\newcommand{\bftab}{\fontseries{b}\selectfont}

\title{Feature salience -- not task-informativeness -- \\drives machine learning model explanations}
\author{Benedict Clark, Marta Oliveira, Rick Wilming, Stefan Haufe}
\date{\today}

\begin{document}

\maketitle

\begin{abstract}
Explainable AI (XAI) promises to provide insight into machine learning models' decision processes, where one goal is to identify failures such as shortcut learning. 
This promise relies on the field's assumption that input features marked as important by an XAI must contain information about the target variable.
However, it is unclear whether informativeness is indeed the main driver of importance attribution in practice, or if other data properties such as statistical suppression, novelty at test-time, or high feature salience substantially contribute.
To clarify this, we trained deep learning models on three variants of a binary image classification task, in which translucent watermarks are either absent, act as class-dependent confounds, or represent class-independent noise.
Results for five popular attribution methods show substantially elevated relative importance in watermarked areas (RIW) for all models regardless of the training setting ($R^2 \geq .45$). By contrast, whether the presence of watermarks is class-dependent or not only has a marginal effect on RIW ($R^2 \leq .03$), despite a clear impact impact on model performance and generalisation ability. 
XAI methods show similar behaviour to model-agnostic edge detection filters and attribute substantially less importance to watermarks when bright image intensities are encoded by smaller instead of larger feature values. 
These results indicate that importance attribution is most strongly driven by the salience of image structures at test time rather than statistical associations learned by machine learning models. 
Previous studies demonstrating successful XAI application should be reevaluated with respect to a possibly spurious concurrency of feature salience and informativeness, and workflows using feature attribution methods as building blocks should be scrutinised. 
\end{abstract}


\section{Introduction}
While Machine Learning (ML) increasingly solves complex problems, trained models often display undesirable behaviour upon deployment.
Models may unduly rely on protected attributes \citep{dressel2018accuracy,clarkcorrecting}.
or rely on confounding relationships between in- and outputs, such as chest tubes and portable radiographic markers, rather than actual disease biomarkers \citep{bellamy2022structural, banerjee2023shortcuts}, thereby bypassing the direct causal relationships they might be expected to learn.
Such reliance on confounding, often termed \emph{shortcut learning} \citep[e.g.,][]{brown2023detecting}, can lead to poor generalisation when these associations shift or vanish in test data \citep{degrave2021ai,brown2023detecting}.

Diagnosing and correcting such failure modes is one of the primary goals of explainable artificial intelligence \citep[XAI,][]{samek2019towardsexplainable}.
Feature attribution methods are XAI tools that assign importance scores to individual input dimensions. It is frequently assumed that these scores reveal `what models have learned', which could then be used to inform subsequent scientific follow-up experiments or to diagnose and possibly improve models, training data, or test inputs. 
In a prominent example, Layer-wise Relevance Propagation (LRP) was demonstrated to highlight a photographer's watermark in a horse vs.\ car classification task \citep{lapuschkinUnmaskingCleverHans2019}. As the watermark was unevenly distributed across classes, it acted as a confounder; the authors attribute this discovery to the method's ability to identify the model's reliance on the artifact. Similar workflows are used in medical imaging to verify if models track pathology rather than confounding artifacts \citep{arias2024analysis, vasquez2025detecting}.

Formally, \emph{confounding} in machine learning occurs when an unobserved variable $C$ causally influences both an input feature $X_i$ and the target $Y$, inducing a statistical association that makes $X_i$ \emph{informative} with respect to $Y$ \citep{pearl2009causality}. 
Conversely, \emph{suppression} occurs when an input feature $X_\text{coll}$ (a \emph{collider}) is caused by both the target $Y$ and a feature $X_\text{supp}$ unrelated to $Y$ (a \emph{suppressor}).
While suppressors are statistically independent of the target (and thus \emph{non-informative}), they introduce non-target-related variance to the collider.
As a result, optimal ML models \emph{must} utilise suppressors to denoise informative features \citep{congerRevisedDefinitionSuppressor1974,haufeInterpretationWeightVectors2014,weichwald2015causal}.
In image classification, a non-informative watermark overlapping with a target object could act as a suppressor, prompting an optimal model to use the watermark to remove it from the object.

Critically, these causal structures have distinct implications: unstable confounding can hurt generalisation, whereas suppression often aids model performance. 
Consequently, \citet{wilmingScrutinizingXAIUsing2022} formulate the presence of a statistical association between feature and target as a necessary condition for feature importance. Theoretical and empirical evidence suggests, though, that popular attribution methods violate this condition by systematically assigning importance to non-informative suppressors \citep{haufeInterpretationWeightVectors2014,kindermansLearningHowExplain2017,wilmingScrutinizingXAIUsing2022,wilmingTheoreticalBehavior2023}, a finding recently extended to non-linear settings \citep{clark2024tetris,oliveiraPretraining2024,wilming2025gecobench}.

It is currently unclear whether task-informativeness can, nonetheless, be considered the main driver of feature importance in practice.   
We hypothesise that suppressor features, novel features not contained in the training data, and features with highly salient structure (e.g., watermarks, sharp edges), might also receive  substantial importance attribution.
To test this, we conduct two controlled studies using local watermarks insertions and global lightness manipulations. 
We evaluate attribution behaviour across three experimental settings: a \emph{confounded setting} where the manipulation is class-specific; a \emph{balanced setting} where the manipulation is non-informative (potentially inducing suppression effects); and a \emph{baseline setting} where manipulations are absent during training but present at test time. By contrasting these settings, we isolate whether attributions really respond to what models have learned (i.e., confounding or suppression), to novel features, or simply to the low-level properties of the test image itself.

In both studies, we train models on datasets generated in all three settings and evaluate the trained models on images with and without experimental manipulations applied.
We measure the relative importance attributed to the watermarked area compared to the entire image (RIW) and the relative importance attributed to the lightness channel compared to all three channels of an HLS colour model (RIL).
With these metrics, we investigate which properties of data and models are most strongly associated with high importance attribution: discriminative information, non-informative suppression, structures unobserved during training, salient low-level properties of the test image (such as sharp edges within the watermarked area or elevated lightness across the entire image), or none of the above.

\section{Results}

We report on extensive experiments in which we studied the dependence of feature attributions on fundamental data and model properties.
To this end, we created synthetic image data by altering real images via controlled manipulations in two different studies -- \emph{watermarks} and \emph{lightness}.
In each, we contrast three different experimental settings where the manipulated image features possess different statistical properties: a \emph{confounded} settings, where manipulated features are informative with respect to the target, a \emph{balanced} settings, in which the same manipulated features are non-informative (yet, potentially induce suppression), and \emph{baseline/no-watermark} settings, in which again the same features constitute artifacts with respect to the training distribution -- they only occur during model evaluation on test data.
Features differing in such properties require different treatment when the goal is to identify actual failure modes and to take corrective measures. 


\subsection{Watermark study}\label{sec:wm-results}
\begin{figure}[!htbp]
    \centering
    {\sffamily \normalsize A} \includegraphics[width=0.8\textwidth]{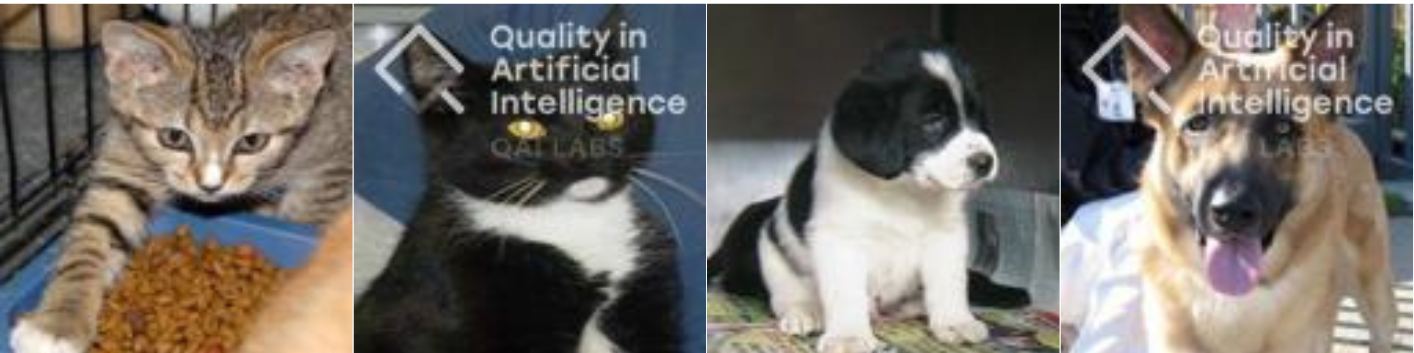}\\
    \vspace{3mm}
    {\sffamily \normalsize B} \includegraphics[width=0.8\textwidth]{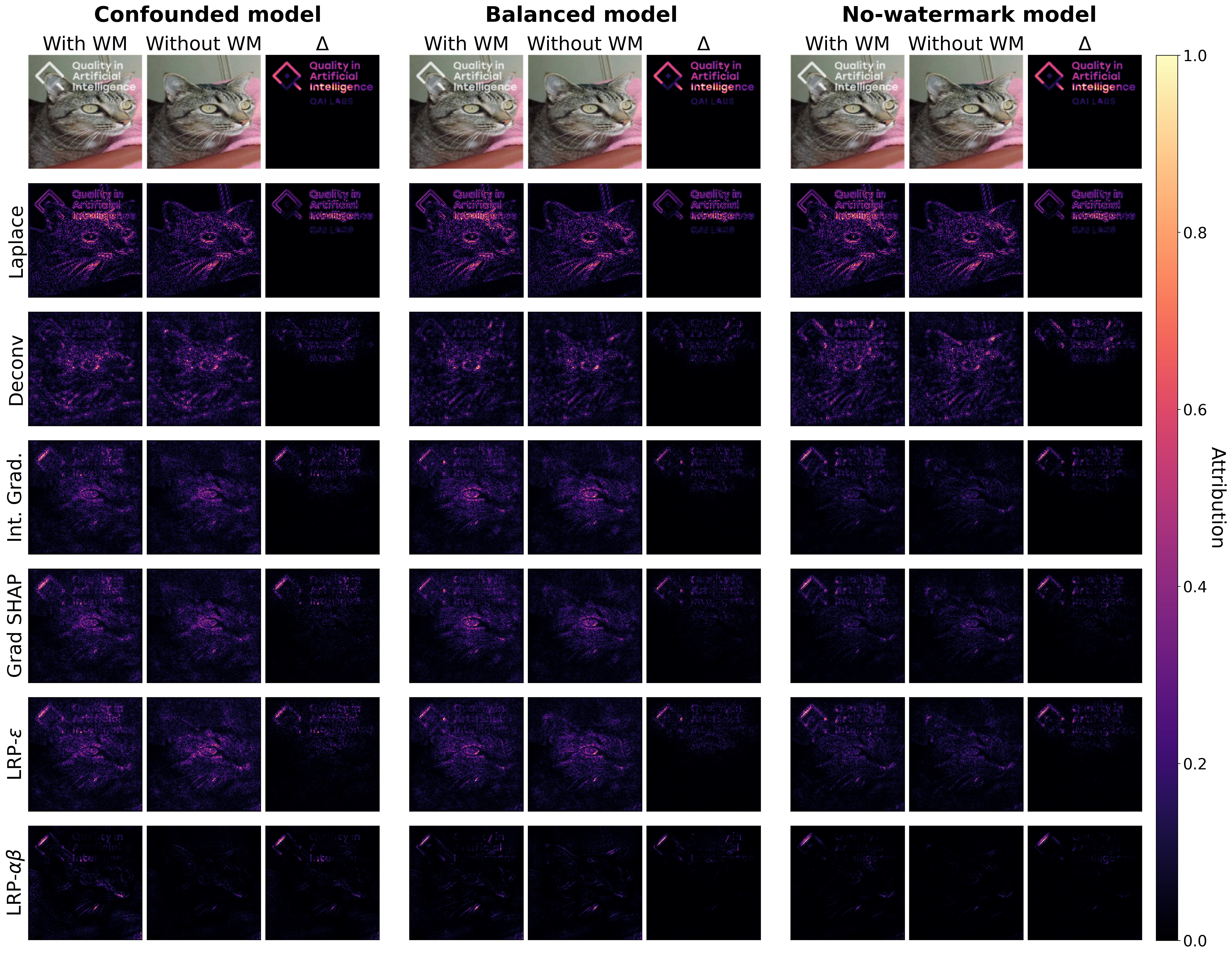}
    \caption{Generated images used and attribution maps obtained in the watermark study. {\sffamily A}: Examples of images of cats and dogs with and without a watermark overlaid at a fixed position. Depending on the experimental setting studied, the watermark's prevalence is either higher for dogs than for cats (confounded setting) or not (balanced setting), or the watermark is not present at all (no-watermark setting) during training. {\sffamily B}: Absolute-valued importance attributed to each pixel of a given test image by different feature attribution methods (Deconvolution, Integrated Gradients, Gradient SHAP, LRP-$\epsilon$, and LRP-$\alpha \beta$) and a model-independent edge-detection filter (discrete 2D Laplace filter). Shown are triplets of columns for models trained in the three settings and applied to test images with (WM) or without (No-WM) the watermark present, where the difference ($\Delta$) between the two attributions is also shown. Explanations obtained by feature attributions resemble results obtained by the model-independent discrete 2D Laplace filter and consistently highlight any present watermark irrespective of whether watermarks introduced confounding during model training or were balanced across cat and dog classes. Even for a model trained on data with no watermarks, substantial importance is attributed to watermarks in test images.}
    \label{fig:data-cats-dogs}
\end{figure}

In the \emph{watermark} study, $N_{\text{cat}}=4,800$ images of cats and the same number of dog images were obtained from Kaggle \citep{kaggle2018catsdogs}, forming the baseline, or \emph{no-watermark}, dataset $\mathcal{D}_{\text{no-wm}}$. Next, a static translucent watermark was superimposed in a fixed position in the upper part of 80\% of all dog but only 20\% of all cat images, forming a separate \emph{confounded} dataset $\mathcal{D}_{\text{conf}}$. 
Finally a third \emph{balanced}, dataset $\mathcal{D}_{\text{bal}}$ was created in which the same watermark was superimposed on 50\% of all images regardless of class membership. In this case, watermarked regions may cover both informative content and non-informative background image parts, where non-informative parts may act as suppressors to clear the watermark from informative parts. 
Examples of the images with and without overlaid watermarks are displayed in Figure~\ref{fig:data-cats-dogs}A.
All three datasets were divided into training and test parts according to five different random splits. Further methodological details on data generation are provided in \nameref{sec:data_watermarks}.

\subsubsection{Models trained on confounded data do not generalise under distribution shifts}\label{sec:dist-shifts}

The training splits of the {confounded}, {balanced}, and {no-watermark} datasets were used to train classification models to distinguish cat from dog images in the corresponding experimental settings.
Convolutional neural network (CNN) models were trained to achieve competitive in-sample performance on the train splits of all three datasets (see \nameref{sec:training} for details).
Five models with different random intialisations were trained per experimental setting.
We then tested each model on the test splits of again all three datasets, resulting in a 3 $\times$ 3 table of area under the receiver operating characteristics (AUROC) classification performance scores (Table \ref{tab:AUROCs}).

In the confounded case, watermarks carry additional discriminative information about the presence of cats and dogs that image classifiers can use.
However, a model relying on watermark information is misled when applied to data in which this information is not present or not discriminative.
Such a confounded model may systematically misclassify images in this case -- being overconfident in predicting `dog' when a watermark is present and `cat' when it is not present.
Thus, we expect that models trained on confounded data should perform better than the other two models when also evaluated on confounded data but worse when evaluated on balanced or no-watermark data. 
Conversely, models trained on no-watermark and suppressor data are both expected to perform similarly well in all test settings, where we might expect a slight advantage of the model trained on balanced data, as this model is exposed to training data both with and without watermarks and is, hence, better positioned to remove or ignore watermarks.

These expectations are confirmed by the observed AUROC scores shown in Table~\ref{tab:AUROCs}, where standard deviations are taken over 25 combinations of five different random data splits and five different random model initialisations.
High test performance (AUROC $> 0.88$) is obtained for models trained and evaluated \emph{in-sample} on identically distributed data for all three experimental settings. 
With AUROC $= 0.92$, the in-sample performance is substantially higher in the confounded than in the balanced (AUROC $= 0.89$) and no-watermark (AUROC $= 0.90$) settings, though.
However, distribution shifts removing confounding relationships have a profound negative impact on model performance at test-time, dropping to AUROC $= 0.79$ for balanced test data and AUROC $= 0.86$ for no-watermark test data; thus, demonstrating an instance of {shortcut learning}, where the model actively uses class-related information contained in the watermark.
In the reverse case, where models trained on unconfounded (balanced) training data are evaluated under confounding, no substantial drop in performance is observed.  
These results reinforce that, as a prerequisite to assisting humans in `debugging' ML models or their respective training data, feature attribution methods should be able to discern informative (thus, possibly confounded) features from non-informative data features. 


\begin{table}[h!]
\sffamily 
\centering
\begin{tabular}{lrrr}
\toprule
Model     &  \multicolumn{3}{c}{AUROC $\times$ 100 on Test Data} \\ 
 & \multicolumn{1}{c}{Confounded} & \multicolumn{1}{c}{Balanced} & \multicolumn{1}{c}{No-watermark} \\ 
 \midrule
Confounded & {\bftab 92.32} $\pm$ 0.006  & 78.98 $\pm$ 0.027 & 86.32 $\pm$ 0.013 \\
Balanced & 88.68 $\pm$ 0.015 & 88.59 $\pm$ 0.007 & {\bftab 89.24} $\pm$ 0.007 \\
No-watermark & 84.73 $\pm$ 0.033 & 87.09 $\pm$ 0.010 & {\bftab 89.77} $\pm$ 0.010 \\ 
\bottomrule
\end{tabular}
\caption{Area under the receiver operating characteristic curve (AUROC) dog vs. cat classification performance under different statistical image manipulations with fixed-position watermarks. Machine learning models trained over confounded, balanced, and no-watermark images were applied to all three corresponding test sets, respectively. Results are averaged over five trained models for each of five random data splits into training, validation, and test sets, with standard deviations across models and splits shown.}
\label{tab:AUROCs}
\end{table}

\subsubsection{Feature importance is explained by test image characteristics, not trained model}\label{sec:wm-test-statistics}

Next, we evaluated how much relative importance different feature attribution methods assigned to watermarked areas depending on the training setting (confounded, balanced, or no-watermark). 
For each image contained in the test set $\mathcal{D}^{\text{test}}_{\text{no-wm}}$ without watermarks, a watermarked variant was created, yielding a corresponding set $\mathcal{D}^{\text{test}}_{\text{wm}}$.
All trained models were subsequently applied to both variants of each test image.
Following, a range of popular feature attribution methods (Deconvolution, Integrated Gradients, Gradient SHAP, LRP) with parameters set to default values were applied.
LRP was tested both with the $\alpha \beta$-variant and the $\epsilon$ with identical parameters that led to the watermark attribution in \citep{lapuschkinUnmaskingCleverHans2019}.
As two model-independent importance attribution baselines, we also considered the discrete 2D Laplace edge-detection filter \citep{gonzalez2009digital} and the test input image itself.
All attribution maps were rectified by taking the absolute value and averaged across colour channels to reflect general feature `importance' not targeted towards a particular class.
We calculated the relative importance within on watermarked area (RIW) as the average importance attributed to the watermarked region, normalised by the average importance distributed over the entire image. 
Since presence of the watermark is the only difference between both samples, we can assess the influence of the watermark on the attribution by quantifying the importance attributed to the watermark features versus the average importance attributed to features of the sample.
Further details on the application of feature attribution methods are provided in \nameref{sec:attribution}.

Depending on whether models were trained in the confounded, baseline, or no-watermark setting, we might expect the RIW to differ strongly.
We discuss several possible hypotheses and the expected outcomes under each:

\begin{enumerate}[label=H{\arabic*}]
    \item \textbf{Informativeness:} First, following established views expressed in the XAI literature, our main hypothesis is that all considered feature attribution methods are most sensitive to watermarks that are informative and present during training.
    \citet{lapuschkinUnmaskingCleverHans2019} ascribe high importance attribution to a watermark to a confounding effect. 
    Under this hypothesis, we expect the RIW to be substantially higher for models trained on confounded data than for models trained in the non-informative balanced and no-watermark settings, on average.
    For fixed-position watermarks, presence and absence of watermarks encodes equivalent information about the image class under confounding.
    Thus, we would expect similar RIW for images containing vs. not containing the watermarks in the confounded setting.
    
    \item \textbf{Suppression:} Numerous works \citep[e.g.,][]{congerRevisedDefinitionSuppressor1974,haufeInterpretationWeightVectors2014,weichwald2015causal} have pointed out that optimal models \emph{must} systematically use non-informative suppressor features, if present, to remove non-target related variance from informative features, and that features attribution methods reflect that behaviour by systematically attributing importance to suppressors \citep{wilmingScrutinizingXAIUsing2022,wilmingTheoreticalBehavior2023,clark2024tetris}.
    Second, we hypothesise that the RIW in the balanced setting will be substantially higher than in the no-watermark setting, but presumably still slightly lower than in the confounded setting. 
    
    \item \textbf{Outliers:} Third, we alternatively hypothesise that high RIW on test images may predominantly occur when watermarks were unobserved during model training, representing novel or outlier features.
    Here, substantially lower RIW would be expected for non-watermarked than for watermarked test images in the no-watermark setting, whereas low RIW for both watermarked and non-watermarked images would be expected in the confounded and balanced settings, in which models are exposed to watermarks during training.
    
    \item \textbf{Salience:} Fourth, we speculate that importance attribution is strongly affected by low-level structural properties of any given image. 
    Watermarks do constitute highly salient features, so may appear in feature attributions primarily for that reason.
    Under this hypothesis, we would expect substantially larger RIW for watermarked than non-watermarked images, regardless of the considered experimental setting.
    Consequently, we would expect any differences in RIW between settings to be small, if not negligible, compared to differences between watermarked and non-watermarked images.
    
    \item \textbf{Null:} Finally, it is conceivable that none of the hypothesised mechanisms explain the importance attribution patterns in the studied settings, or that multiple of the hypothesised mechanisms are jointly affecting the RIW.
    
\end{enumerate}

Note that only under hypotheses H1 and H2 could it reasonably be argued that feature attributions reflect what an underlying model has learned, and only under hypothesis H1 can feature attributions effectively aid AI quality assurance efforts \citep{haufe2024position}. 

\begin{figure}[ht!]
    \centering
        \includegraphics[width=0.95\textwidth]{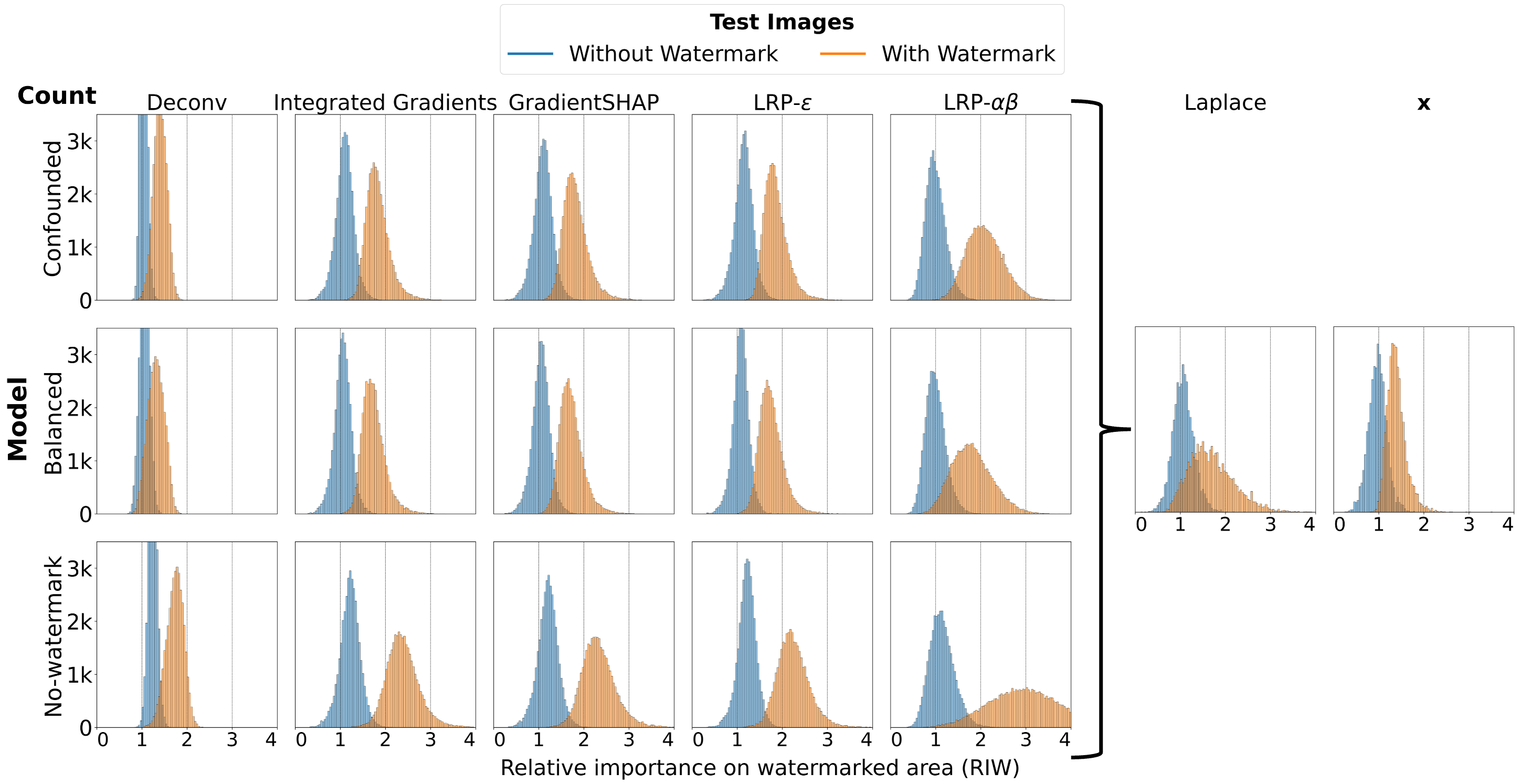}
    \caption{Distributions of the relative importance on watermarks (RIW), quantifying the amount of importance attributed to pixels manipulated by insertion of a fixed-position translucent watermark in the watermark study, relative to the entire image. RIW distributions are shown for identical test images with and without the actual watermark applied, for different machine learning models (shown across rows), for which watermarks acted either as confounders (confounded setting), non-informative features (balanced setting) during training, or were completely absent during training (no-watermark setting). This is shown for a selection of feature attribution methods (Deconvolution, Integrated Gradients, Gradient SHAP, LRP-$\epsilon$, and LRP-$\alpha \beta$) as well as two model-independent baselines (Laplace and raw images $\mathbf{x}$), shown across columns. 
    It can be seen that pixels manipulated by the watermark are attributed substantially higher RIW if a watermark is actually present, compared to the case where no watermark is present. This is true across all studied attribution methods and baselines but also across all experimental settings including the balanced and no-watermark settings, in which the presence of a watermark is non-informative about the class label.
    }
    \label{fig:hist_watermark_fixed}
\end{figure}

By contrast, our results strongly align with hypothesis H4 (Salience, Figure \ref{fig:hist_watermark_fixed}). While we observe large RIW on watermarked images in the confounded setting, confirming the anecdotal observation of \citet{lapuschkinUnmaskingCleverHans2019}, the same is true for both other settings as well.
Across all settings, the observed RIW are, to a large extent, explained by the presence or absence of the watermark.
Meanwhile, the statistical role of watermarks (confounded, suppressor, or outlier) for models trained in different settings has far less influence on RIW for all studied feature attribution methods.
Thus, it appears that RIW is mainly driven by low-level image properties, such as highly salient edges and corners, while what the model learned during training appears to play a negligible role for importance attribution.

Below, we report the experimental results obtained in experiments with fixed-position watermarks, which led to our assessment, while equivalent results for variable position experiments are presented in Supplementary Table~\ref{tab:AUROCs-variable} and Supplementary Figure~\ref{fig:hist_watermark_variable}.
Results obtained in the lightness study, presented in \nameref{sec:coco-results} further corroborate our interpretations.

Figure~\ref{fig:data-cats-dogs}B shows pixel-wise importance attributions obtained by five methods and one model-independent baseline obtained for a single test image of a dog image with and without the watermark applied. 
Regardless of whether the model to be `explained' was trained on confounded, balanced, or no-watermark data, the watermark is highly visible in the attribution maps produced by all five methods.
The watermark increases further in salience, effectively remaining the only salient structure, when attributions obtained for the same model and attribution method and for the same test image without overlaid watermark are subtracted from one another.
In all three settings, explanations obtained by feature attribution methods are qualitatively similar to results obtained by applying the discrete Laplace edge-detection filter to the raw input images.

While results presented Figure~\ref{fig:data-cats-dogs}B pertain to one single image, singular value decomposition (SVD) confirms the watermark as the most salient structure across the attributions of all watermarked test images. 
Supplemental Figure~\ref{fig:explanation-pca} shows the absolute-valued first singular vector of the pixel $\times$ sample attribution matrix, representing the image-space direction explaining the highest amount of attribution variation across the test samples with watermarks in the fixed position.
This is shown for models trained on confounded, balanced, and no-watermark data, respectively, in combination with all five attribution methods and two model-independent baselines.
The watermark is clearly discernible in all cases, often again representing the single most salient structure. 
Moreover, the first singular vectors of the Integrated Gradients, Gradient SHAP, and LRP-$\epsilon$ attribution maps show similarity to the first singular vectors obtained from the Laplace filter image baseline.


Quantitatively, Figure \ref{fig:hist_watermark_fixed} shows histograms of the RIW scores obtained from all images with and without watermarks in the five different test sets, and for five randomly initialised models each, separately for each experimental setting and attribution method.
In all studied cases, the presence of a watermark in test images leads to substantially higher RIW compared to the same images without watermark applied, in line with what is shown qualitatively in Figure~\ref{fig:data-cats-dogs}B.
In all cases, the RIW distributions for watermarked and non-watermarked images are strongly separated, where distributions related to non-watermarked images tend to centre around RIW $=1$, indicating importance values consistent with the global average across images.
Meanwhile, RIW scores for images with watermarks applied tend to centre around RIW $=2$, indicating that twice as much importance is attributed to pixels within the watermarked area than to the average image pixel.
In this, histograms for all feature attribution methods closely resemble those obtained for raw images and Laplacian filters.

In contrast, RIW distributions obtained for any given feature attribution method, for either only watermarked or only non-watermarked test images, differ less strongly across the three studied experimental settings.
Differences for non-watermarked images are almost impossible to discern visually for all methods.
Histograms for watermarked images do show visually noticeable differences, where substantially higher RIW is observed in the no-watermark setting than in the confounded and balanced settings, whereas differences between the latter two are again hard to discern.
An exception is the LRP-$\alpha \beta$ variant, which does attribute higher importance to watermarks in the confounded than in the balanced setting, while the highest importance is attributed in the no-watermarks setting consistent with other methods.

Means and standard errors for all RIW distributions are shown in Supplementary Table~\ref{tab:mean_se}, where an additional distinction between cat and dog images is made.
Notably, there are small but significant RIW differences between cat and dog images even in the balanced and no-watermarks settings for various feature attribution methods as well as for general Laplace-filtered raw images, pointing to differences in the image statistics of the two classes that are independent of our controlled manipulation.
For this reason, we did not further investigate the effect of test image class on RIW. 

For the factors training setting (confounded, balanced, no-watermark) and test image characteristics (watermark present or not), we used linear mixed models (LMM) to quantify the amount of RIW variation explained by either settings or characteristics in terms of $R^2$ values, where we also assessed the statistical significance of differences between pairs of factor levels. Results are presented in Table~\ref{tab:R2}. Across all three studied settings, the presence of a watermark in test images explains between 45\,\% (Deconvolution) and 67\,\% (LRP-$\epsilon$) of the RIW variation, where percentages are obtained as $100 \cdot R^2\,\%$
For any given pair of settings, the range of attained $R^2$ values is similar (between 42\,\% for Deconvolution and 72\,\% for LRP-$\epsilon$. Laplace (39\,\%) and raw image (51\,\%) baselines achieve scores in a similar range. 
In all cases, RIW is significantly higher for watermarked than for non-watermarked test images ($p < 10^{-16}$).

The setting within which models were trained explains only between 8\,\% (LRP-$\epsilon$) and 31\,\% (Deconvolution) of the variation in RIW, when considering the pooled data of all three settings. When considering confounded and balanced settings only, the setting only explains between 2\,\% (Deconvolution, Integrated Gradients, Gradient SHAP) and 3\,\% (LRP variants) of the RIW variation, indicating that all studied methods possess next to no ability to statistically discriminate between watermarks representing informative features and those representing non-informative features for the underlying model. Despite the small effect sizes, all methods, however, consistently attain higher RIW in the confounded setting ($p < 10^{-16}$).

The no-watermark setting was confirmed to yield substantially higher RIW than the confounded and balanced settings for all attribution methods with setting explaining between 3\,\% (LRP-$\epsilon$) and 26\,\% (Deconvolution) of the RIW variation when comparing confounded and no-watermark models, and between 10\,\% (LRP-$\epsilon$) and 33\,\% (Deconvolution) when comparing balanced and no-watermark models. Again, all observed effects are statistically significant with $p < 10^{-16}$.

\begin{table}[h!]
    \sffamily
    \centering
    \begin{small}
    \tabcolsep2mm
    \begin{tabular}{lrrrrrp{1mm}rr}
    \multicolumn{9}{l}{\textbf{A}: Percent variance of relative importance on watermarked area (RIW) explained by watermark presence}\vspace{0.5em}\\
        \toprule
        Model training settings 
        & \multicolumn{1}{c}{Deconv} & \multicolumn{1}{c}{IG} & \multicolumn{1}{c}{GradSHAP} & \multicolumn{1}{c}{LRP-\(\epsilon\)} & \multicolumn{1}{c}{LRP-\(\alpha\beta\)} & & \multicolumn{1}{c}{Laplace} & \multicolumn{1}{c}{\textbf{x}}\\
        & \multicolumn{8}{c}{$R^2 \, \times$ 100 [\%], $\uparrow$ denotes higher RIW in watermarked test images} \rule{0pt}{3ex}\\
        \midrule
        \{Confounded, Balanced, No-WM\} & $\uparrow$ 45 & $\uparrow$ 64 & $\uparrow$ 63 & $\uparrow$ 67 & $\uparrow$ 54 & \multirow{4}{*}{\resizebox{2.5mm}{8mm}{$\Biggr\}$}} & \multirow{4}{*}{$\uparrow$ 39} & \multirow{4}{*}{$\uparrow$ 51} \\
        \{Confounded, Balanced\} & $\uparrow$ 61 & $\uparrow$ 70 & $\uparrow$ 68 & $\uparrow$ 70 & $\uparrow$ 67 &  & & \\
        \{Confounded, No-WM\} & $\uparrow$ 54 & $\uparrow$ 70 & $\uparrow$ 69 & $\uparrow$ 72 & $\uparrow$ 63 & & & \\
        \{Balanced, No-WM\} & $\uparrow$ 42 & $\uparrow$ 62 & $\uparrow$ 62 & $\uparrow$ 65 & $\uparrow$ 50 &  & & \\
    \bottomrule
    \vspace{0.75em}\\
    %
         \multicolumn{9}{l}{\textbf{B}: Percent variance of relative importance on watermarked area (RIW) explained by model training setting} \vspace{0.5em}\\
        \toprule
        Model training settings  & \multicolumn{1}{c}{Deconv} & \multicolumn{1}{c}{IG} & \multicolumn{1}{c}{GradSHAP} & \multicolumn{1}{c}{LRP-\(\epsilon\)} & \multicolumn{1}{c}{LRP-\(\alpha\beta\)} & & \multicolumn{1}{c}{Laplace} & \multicolumn{1}{c}{\textbf{x}}\\
        & \multicolumn{8}{c}{$R^2 \, \times$ 100 [\%], $\leftarrow$ denotes higher RIW in setting A for settings \{A,B\} }  \rule{0pt}{3ex}\\
        \midrule
        \{Confounded, Balanced, No-WM\} & 31 & 11 & 10 & 8 & 15 & \multirow{4}{*}{\resizebox{2.5mm}{8mm}{$\Biggr\}$}} & \multirow{4}{*}{0} & \multirow{4}{*}{0 } \\
        \{Confounded, Balanced\} & $\leftarrow$ \hphantom{0}2 & $\leftarrow$ \hphantom{0}2 & $\leftarrow$ \hphantom{0}2 & $\leftarrow$ \hphantom{0}3 & $\leftarrow$ \hphantom{0}3 &  & \\
        \{Confounded, No-WM\} & $\rightarrow$ 26 & $\rightarrow$ \hphantom{0}7 & $\rightarrow$ \hphantom{0}6 & $\rightarrow$ \hphantom{0}3 & $\rightarrow$ 10 & & \\
        \{Balanced, No-WM\} & $\rightarrow$ 33 & $\rightarrow$ 13 & $\rightarrow$ 13 & $\rightarrow$ 10 & $\rightarrow$ 17 &  & \\
    \bottomrule
    \end{tabular}
    
    \end{small}
    \caption{Variance of relative importance on watermarked area (RIW) scores explained by, {\sffamily A}, whether a watermark is present in a given test image or not, and, {\sffamily B}, experimental setting, whereby machine learning models are either trained on data confounded by the occurence of watermarks (confounded setting), data with equal prevalence of watermarks in cat and dog image classes (balanced setting), or data without any watermarks (no-watermark setting). Variance explained is measured as the coefficient of determination ($R^2$) of the linear mixed models (LMM) RIW $\sim$ WM and RIW $\sim$ setting. LMMs were fitted separately for different attribution methods (Deconvolution, Integrated Gradients, Gradient SHAP, LRP-$\epsilon$, and LRP-$\alpha \beta$) as well as model-independent baselines (Laplace and raw images $\mathbf{x}$), and for different combinations of experimental training settings (all settings as well as all pairwise combinations of two settings) on data of both image classes (cats and dogs). The presence of a watermark in a test image leads to significantly higher RIW in all studied cases, as marked by upward pointing arrows, where $0.42 \leq R^2 \geq 0.70$ and $p < 10^{-16}$ in all cases. By contrast, whether watermarks were class-informative (confounded setting) or completely non-informative with respect to the class label (balanced setting) during training explained no more than three percent of the RIW variation ($R^2 \leq 0.03$) for any studied attribution method, with marginally yet significantly ($p < 10^{-16}$) higher RIW in the confounded than the balanced setting, indicated by leftward pointing arrows. Models trained on data without watermarks (no-watermark setting) produced significantly higher importance on watermarked test images than models trained in the confounded and baseline settings for all attribution methods studied ($0.03 \leq R^2 \geq 0.33$, $p < 10^{-16}$), as indicated by rightward pointing arrows. Note that Laplace and raw-image baselines are model-independent and therefore identical in the three studied settings.}
    \label{tab:R2}
\end{table}

\subsubsection{Inverted colour encoding reverses feature importance attribution}
Watermarks, as applied here, lead to a general brightening of the watermarked area as seen in Figure~\ref{fig:data-cats-dogs}A, which is comparable to the situation studied in \citet{lapuschkinUnmaskingCleverHans2019}. In the standard RGB colour encoding we used, where 1 encodes brightest and 0 encodes darkest colours, this leads to an upward shift in pixel intensities in the watermarked area and a corresponding upward shift in the RIW scores for watermarked images, as seen in Figure~\ref{fig:hist_watermark_fixed}. At the same time, watermarks introduce salient image structure through sharp edges, which are picked up by spatial derivative operators, leading to an upward shift also for the RIW of the discrete Laplacian.

In what has been reported so far, all studied feature attribution methods display a substantial upward shift for watermarked images regardless of the training setting of the underlying model, suggesting that these methods primarily reflect differences in local pixel intensities, local pixel gradients, or both, consistent with hypothesis H4. To distinguish between these three possibilities, we repeat our experiments for images with inverted colour encoding, where 1 now encodes darkest and 0 encodes brightest colours. Notably, this transformation leaves the image itself entirely intact. Likewise, the (absolute-valued) discrete Laplacian studied here is invariant to this reversal in colour encoding. 

Supplemental Table~\ref{tab:AUROCs-inverted} presents AUROC classification accuracies obtained in the inverted colour encoding experiment, which are qualitatively and quantitatively equivalent to the results obtained for standard encoding (Tables~\ref{tab:AUROCs} and \ref{tab:AUROCs-variable}). 
Corresponding RIW distributions are shown in Supplementary Figure~\ref{fig:hist_watermark_inverted}, where RIW scores of raw pixel intensities now generally display a downward shift for watermarked images, while scores obtained from discrete Laplacian images show the familiar upward shift. Interestingly, feature attribution methods do not show consistent behaviour in this scenario. Deconvolution, and LRP-$\alpha\beta$ show the familiar RIW upward shift for watermarked images in all three studied settings, which is the reverse behaviour seen for raw pixel intensities but consistent with the behaviour seen for the Laplacian. This suggests that these two methods are most sensitive to sharp edges and other salient structures in test images than to raw pixel intensities. Conversely, GradientSHAP and LRP-$\epsilon$ qualitatively display a noticeable RIW downward shift for watermarked images in all studied settings, and Integrated Gradients shows a downward shift for the confounded and balanced settings, whereas no visually noticeable shift is observed in the no-watermark setting. This suggests that GradientSHAP and LRP-$\epsilon$ are highly sensitive to raw feature values. 

\subsection{Lightness study}\label{sec:coco-results}
\begin{figure}[!ht]
    \centering
       \includegraphics[width=0.9\textwidth]{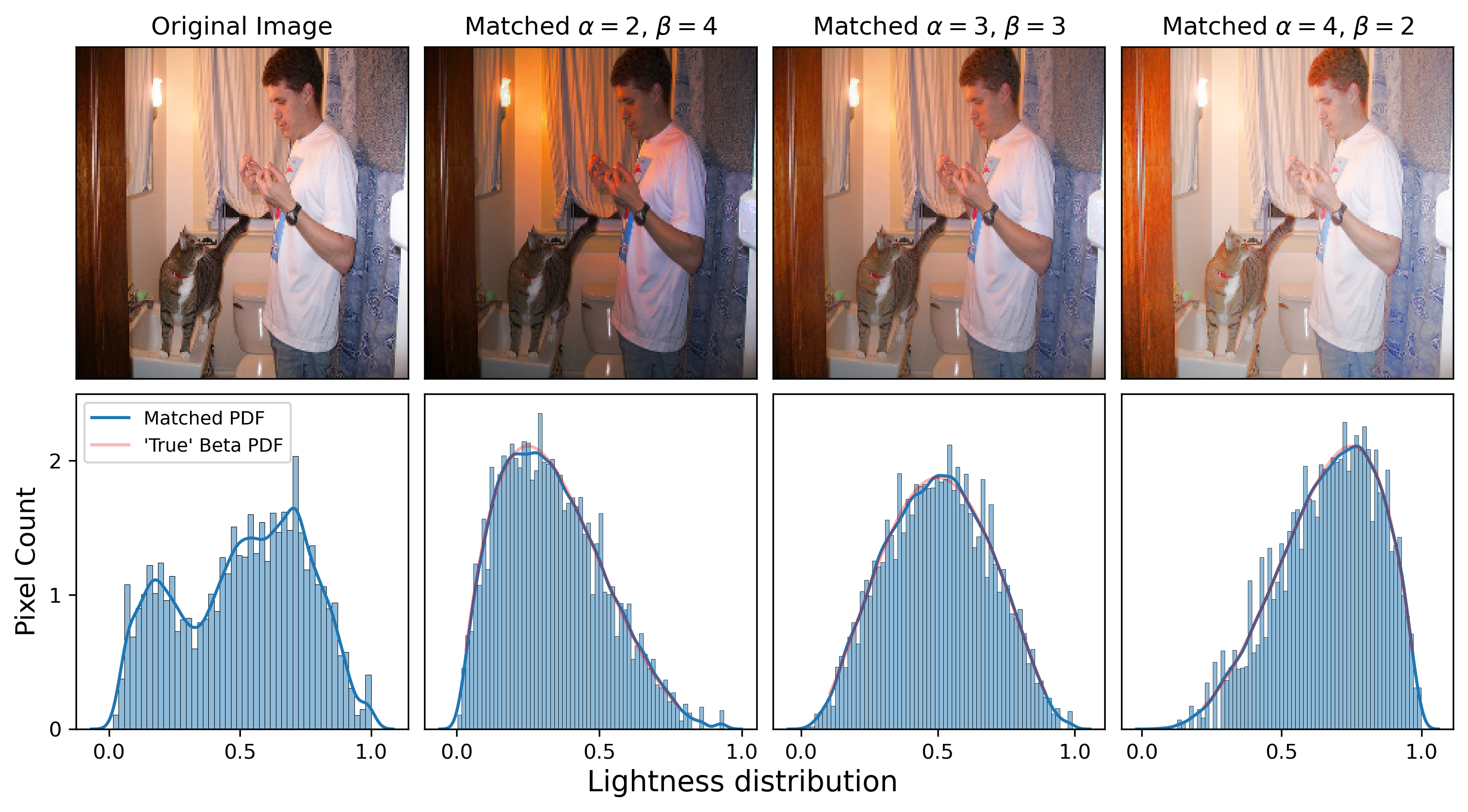}
    \caption{Examples of different lightness manipulations applied to the lightness channel of an animal image (upper row) in the lightness study, where the histogram of the lightness distribution is transformed to match a given beta distribution (lower row). Beta distributions $\text{Beta}(2,4)$ and $\text{Beta}(4, 2)$ were imposed on 50\% of the images either dependent (confounded setting) or independent (balanced setting) of membership in one of two studied classes (animals and vehicles). In all non-manipulated images as well as all images in the baseline setting, lightness levels are normalised to a centred $\text{Beta}(3,3)$ distribution. }
    \label{fig:data-coco}
\end{figure}

For the lightness study, $N_{\text{animal}} = N_{\text{vehicle}} = 15,000$ images of animals and vehicles were sourced from the Common Objects in Context \citep[COCO,][]{tsung2014microsoftcoco} dataset. Images were converted from RGB to HLS (Hue, Lightness, Saturation) colour space with all values normalised to range $[0, 1]$. The lightness channel of all images was further normalised to follow a central Beta(3,3) distribution, giving rise to the \emph{baseline} dataset $\mathcal{D}_{\text{base}}$. Similar to the watermark study, manipulations were applied to again yield a confounded dataset  $\mathcal{D}_{\text{conf}}$ and a balanced dataset  $\mathcal{D}_{\text{bal}}$. In contrast to the watermark study, though, manipulations were not local but affected the lightness of the entire image, where, in the confounded setting, 
50\% of the images of the animal class were darkened and 50\% of the images of the vehicle class were lightened by imposing $\text{Beta}(2,4)$ and $\text{Beta}(4,2)$ distributions on the lightness channel, respectively. In the balanced setting, 25\% of the images of both classes were darkened and 25\% were lightened. Figure~\ref{fig:data-coco} depicts a raw image from COCO with the three different lightness manipulations (darkened, normalised, brightened) applied.

Image classifiers trained and evaluated on either baseline, confounded, or balanced data achieved classification accuracies of AUROC $> 0.89$. When transferring models trained in the confounded setting to other settings, a sharp drop in performance was observed. No comparable drop was observed when applying models trained in the baseline or balanced settings. These results, which are consistent with what has been reported for the watermark study, are presented in Table~\ref{tab:AUROCs-coco}.

\begin{table}[h!]
\sffamily 
\centering
\begin{tabular}{lrrr}
\toprule
Model     &  \multicolumn{3}{c}{AUROC $\times$ 100 on Test Data} \\ 
 & \multicolumn{1}{c}{Confounded} & \multicolumn{1}{c}{Balanced} & \multicolumn{1}{c}{Baseline} \\ 
 \midrule
Confounded & {\bftab 96.77} $\pm$ 0.002  & 64.29 $\pm$ 0.014 & 87.02 $\pm$ 0.009 \\
Balanced & 89.18 $\pm$ 0.008 & 89.14 $\pm$ 0.007 & {\bftab 89.55} $\pm$ 0.007 \\
Baseline & 87.51 $\pm$ 0.021 & 87.71 $\pm$ 0.008 & {\bftab 89.83} $\pm$ 0.007 \\ 
\bottomrule
\end{tabular}
\caption{Area under the receiver operating characteristic curve (AUROC) animal vs. vehicle image classification performance under different statistical lightness manipulations. Machine learning models trained over confounded, balanced, and baseline images were applied to all three corresponding test sets, respectively. Results are averaged over five trained models for each of five random data splits into training, validation, and test sets, with standard deviations across models and splits shown.}
\label{tab:AUROCs-coco}
\end{table}

\begin{figure}[!ht]
    \centering
    \includegraphics[width=1\textwidth]{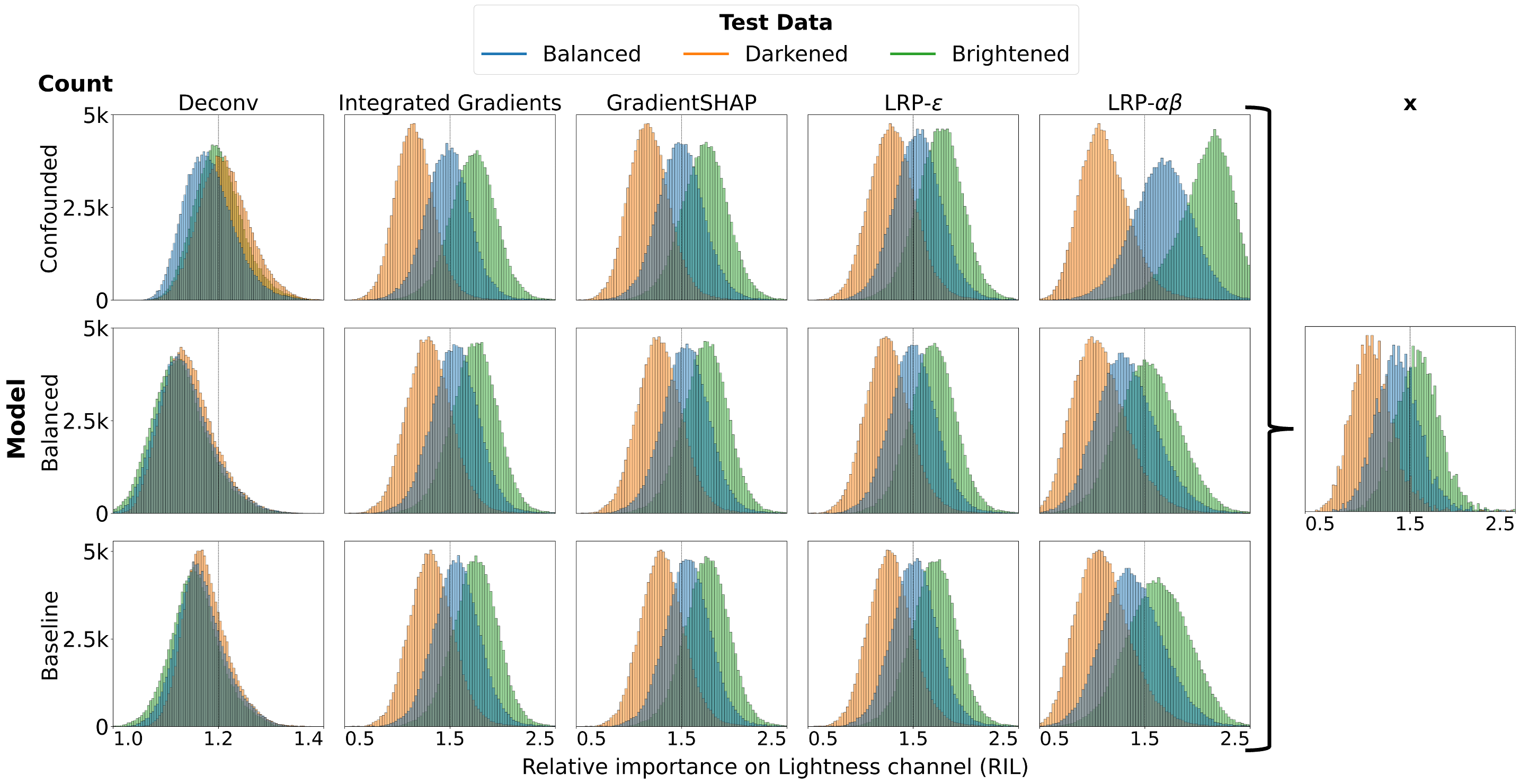}
    \caption{Distributions of the relative importance on lightness (RIL), quantifying the amount of importance attributed to the lightness relative to all three channels of HLS-encoded animal and vehicle images in the lightness study. RIL distributions are shown for identical test images with lightness channels either normalised to a Beta$(3,3)$ distribution, darkened according to a Beta$(2,4)$ distribution or brightened according to a Beta$(4,2)$ distribution, for different machine learning models (shown across rows), for which lightness manipulations either induced confounding (confounded setting), were applied independent of class membership (balanced setting) during training, or were completely absent during training (baseline setting), and for a selection of feature attribution methods (Deconvolution, Integrated Gradients, Gradient SHAP, LRP-$\epsilon$, and LRP-$\alpha \beta$) as well as a model-independent baseline (raw images $\mathbf{x}$), shown across columns. 
    The importance attributed to the lightness channel is strongly determined by the lightness of the test images with brightened images receiving higher importance than normalised baseline images, and baseline images receiving higher importance than darkened images. In contrast, there are no notable changes in attribution distributions for models trained in the confounded, balanced, or baseline settings.}
    \label{fig:hist_lightness}
\end{figure}

Figure \ref{fig:hist_lightness} shows histograms of the proportion of importance attributed to the lightness channel (RIL) on test images with each of the three different lightness manipulations applied ($\mathcal{D}^{\text{test}}_{\text{base}}$, $\mathcal{D}^{\text{test}}_{\text{dark}}$, and $\mathcal{D}^{\text{test}}_{\text{bright}}$ datasets) for each combination of model training setting and feature attribution. Global lightness manipulations are clearly noticeable as up- and downward shifts of raw image intensity $\mathbf{x}$, where darker images show lower RIL scores and brighter images show higher scores. The same behaviour is seen for all feature attribution methods except Deconvolution, suggesting that these methods assign importance primarily based on pixel intensity. In contrast, Deconvolution displays no noticeable differences in RIL distributions between lightness manipulations in the balanced and baseline settings, and shows higher RIL for both darkened and brightened images than for baseline images in the confounded setting.


\section{Discussion}

When feature attribution methods are used to `explain' machine learning model decisions, it is common to assume that strongly attributed features are informative for the prediction task. In an image classification task, for example, the highlighted image parts would be expected to contain objects whose probability of occurrence differs between classes. Under this assumption, which can be formalised by the statistical association property \citep[SAP,][]{wilmingScrutinizingXAIUsing2022,wilmingTheoreticalBehavior2023,haufe2024position}, feature attribution methods might support downstream goals such as model and data diagnostics and correction, scientific discovery, and algorithmic recourse \citep{haufe2024position}.

The belief that popular feature attribution methods like LIME \citep{ribeiroWhyShouldTrust2016}, LRP \citep{binderLayerWiseRelevancePropagation2016}, Integrated Gradients \citep{sundararajanAxiomaticAttributionDeep2017} and SHAP \citep{lundbergUnifiedApproachInterpreting2017} possess the SAP is widespread both among the developers and many users of these methods. Thus, when \citet{lapuschkinUnmaskingCleverHans2019} observe a photographer's watermark in an LRP attribution map for a car vs. horse image classification task, and subsequently identify this watermark to be class specific through confounding, these two concomitant observations are combined, forming the powerful narrative that LRP can systematically reveal what a model has learned, and can thereby aid the detection of shortcut learning on confounded features in practice.

\subsection*{Feature salience dominates importance attribution independent of what models learned}
We conducted the first controlled study to systematically scrutinise this narrative (corresponding to the informativeness hypothesis H1) against suitable alternative hypotheses and null models such as predominant attribution to suppressors (H2), predominant attribution to outlier features (H3), and predominant attribution to salient features independent of an underlying model (H4). Our results speak strongly in favour of H4: salience. 

In the watermark study, we observe high relative importance on watermarks (RIW $\sim 2$) for all feature attribution methods, regardless of the training setting (confounded, balanced, or no-watermark). Conversely, the same areas receive substantially lower importance (close to the image-wide average, RIW $\sim 1$) in all settings for test images that do not contain the watermark. This result is observed both for fixed and variable-position watermarks. Under the informativeness hypothesis (H1), we could have expected the difference between watermarked and non-watermarked images to be less pronounced in the fixed-position case. 
In this setting, the specific spatial location of the watermark is a fixed proxy for the target; thus, both the presence or absence of the watermark provide equivalent discriminative information. 
In the variable-position case, no specific spatial prior exists, implying that information regarding the watermark's absence is distributed globally across the image.

Qualitatively similar results are obtained in the lightness study.
Models attribute higher importance on the lightness channel for globally brightened images and lower importance for globally darkened images -- regardless of whether these lightness manipulations are dependent on the image class, balanced across classes, or not present during training.
Most notably, models trained on non-manipulated data yield RIW and RIL scores comparable to those obtained for models for which these manipulations are either task-informative (confounded) or represent nuisance signals that need to suppressed by the model (balanced).
In light of these findings, it appears invalid to maintain the perspective that attributions primarily reflect model behaviour. Instead, it appears that all tested attributions are strongly dominated by low-level properties of the test images themselves, largely independent of the actual model.

The fact that similar feature attribution behaviour is observed in all experimental settings suggests that a feature's attribution is neither strongly driven by being informative (e.g., confounded), nor by representing a suppressor or outlier.
These observations rule out hypotheses H1--3, leaving only hypotheses H4 (feature salience) and H5 (any other uncontrolled properties) to explain feature attribution behaviour. 
Of these, the following observations strongly favour feature salience (e.g., high intensity or high local contrast) as the main driver of feature importance for the studied combination of data, models, and feature attribution methods in the context of standard natural image prediction pipelines.
Darkening manipulations lead to substantially lower importance attributions to lightness variables than brightening manipulations in all studied settings and for all considered feature attribution methods except Deconvolution.
This is despite the fact that both darkening and brightening manipulations occur with identical frequencies and statistical properties in the training data.
Similarly, with the exception of LRP-$\alpha \beta$ and again Deconvolution, we observe that the exact same watermarked images receive substantially \emph{lower} (instead of higher) relative importance on the watermarked area than their non-watermarked counterparts when the colour encoding is inverted, such that darker instead of brighter pixels are encoded by higher numbers. 
We also observe that most tested attribution methods closely reproduce the behaviour of a simple model-independent edge detection filter like the 2D Laplacian, or even the intensity of the image itself, an observation also made in
\citet{clark2024tetris}.
Based on the inverted colour encoding results, we can further infer that it is primarily 
raw intensity rather than local spatial gradients measured by the rectified Laplacian that most strongly correlates with importance attributions. This is evidenced by the fact that inverted colour encoding reverses relative importance on watermarked areas, while leaving the (absolute-valued) Laplacian unchanged. 

High salience of informative image structures may indeed play a key role in numerous seemingly convincing applications of feature attribution methods beyond the scenarios studied here.
In Atari video games \citep{lapuschkinUnmaskingCleverHans2019}, informative structures often contrast sharply with uniform backgrounds.
Similar considerations could be made for several real-world applications, for example in medicine. 
In magnetic resonance imaging, brain lesions often visually stand out against the surrounding tissue \citep{hofmann2022towards,martin2026investigating}, as do cancerous cells in histopathology images \citep{klauschen2024toward}. Successful detection of such objects may reflect biases towards salient structures rather than task-informativeness. 
To prevent that XAI methods appear right for the wrong reasons, existing studies demonstrating successful XAI application should be reevaluated with respect to a possibly spurious concurrency of feature salience and informativeness, and workflows using feature attribution methods as building blocks should be scrutinised. 

Similar considerations also apply to research using ground-truth data (obtained either annotation or controlled image manipulation) to benchmark feature attribution methods. Ground-truth structures that stand out visually, such as artificial lesions inserted into medical images, may upwardly bias explanation performance evaluations.
This calls for novel benchmarks, in which informative features are normalised with respect to local and global image statistics and blend into the background more subtly.
\citet{siegel2025explainable} further discuss how differences in image statistics between brain MRI and natural images affect the quality of different feature attribution methods. 

\subsection*{Consistency and differences across feature attribution methods}
The observed results are qualitatively equivalent and quantitatively comparable across all studied feature attribution methods apart from a few notable exceptions. 

In the lightness study, Deconvolution attributions do not seem to show a dependency on the lightness manipulation in the balanced and baseline settings of the lightness study. In the confounded setting, however, darkness and lightness manipulations both lead to slightly but significantly higher relative importance on lightness features for deconvolution. Thus, this method appears to show desired behaviour consistent with the experimentally induced statistical properties of the data.
For all other methods, darkness and lightness manipulations consistently lead to lower and higher RIL compared to baseline images across all studied settings, respectively. 

No such differences between methods are observed in the watermark study with standard colour encoding. Interestingly, when inverting the colour encoding, the behaviour of most feature attribution methods also reverses in the sense that the presence of watermarks now leads to lower instead of higher RIW scores, coincident with lower pixel intensities.
Exceptions are LRP-$\alpha \beta$ and Deconvolution, which both attribute higher RIW to watermarked images as under the original colour encoding. These effects are, however, not constrained to the confounded setting, but are also observed in the no-watermark settings for LRP-$\alpha \beta$, and across all studied settings for Deconvolution.
Strikingly, LRP-$\epsilon$ and LRP-$\alpha \beta$ show opposing behaviour in this experiment.

Despite qualitatively similar results observed across feature attribution methods and experimental settings, certain quantitative differences can be observed.
The training setting explains up to 33\% of RIW variability when comparing the no-watermark setting to the other two. 
Notably, all settings statistically differ from another in terms of RIW, although differences between confounded and balanced settings are very small, explaining at most 3\% of the RIW variability.
In the lightness study, watermarks in test images lead to a wider separation of the RIL for darkened, baseline, and brightened images in the confounded than the other two training settings, an effect that is particularly pronounced for LRP-$\alpha \beta$. 
These differences between settings suggest that feature salience, while being the dominant factor driving importance attribution, is not the only factor.
Both task-informativeness and absence of a feature during training can further contribute to the importance attributed to a feature depending on task characteristics and the feature attribution method used.
This indicates that hypotheses H1 and H3 cannot be \emph{completely} dismissed.

\subsection*{Explainable AI and causal structure in data}

Shortcut learning in ML models can occur if a feature and the target are confounded; that is, if both are effects of a common unobserved cause, as in the watermark example discussed throughout the paper. Confounding can be contrasted to the situation where a feature $X_i$ exerts a genuine causal influence $X_i \rightarrow Y$ on the target $Y$, as well as to the case when a feature is genuinely caused by the target, $Y \rightarrow X_i$. A causal relation from feature to target would, for example, be present if a feature $X_i$ encodes the presence of a genetic mutation that increases the likelihood of developing a disease $Y$. Conversely, an anti-causal relation would be present if a target disease $Y$ affects a certain anatomical structure $X_i$ in a medical image $X$. All three causal structures would lead to a statistical association between $X_i$ and $Y$ that could be exploited by an ML model for prediction in equal measure. As such, ML models are typically neither intended nor capable to distinguish between causal, anti-causal and confounded feature-target associations.
The same holds for feature attribution methods, which are typically applied post-hoc on fitted ML models. 

Thus, when studies apply feature attribution as part of a pipeline designed to detect shortcut learning \citet{lapuschkinUnmaskingCleverHans2019,degrave2021ai,andersRemovingHans2022}, the goal is mainly to screen for features that are statistically associated with the target. The decision whether a given feature is indeed confounded or rather genuinely causal (or anti-causal) with respect to the target is, thereby, left to human domain experts.
Research has, however, shown that feature attribution methods are not designed to detect informative features in the first place but also systematically assign importance to suppressors not statistically associated with the target \citep{haufeInterpretationWeightVectors2014,wilmingTheoreticalBehavior2023}. With this, their utility as screening tools as discussed above may be disputed \citep{haufe2024position}.

Distinguishing features with different causal roles (informative versus suppressor features, for example) requires methods of causal inference that currently go beyond the capabilities of XAI methods.
A complete identification of the causal structure of a dataset would, however, be possible only under strong limiting assumptions \citep{scholkopfCausalityMachineLearning2019}.
Nevertheless, current techniques can answer a variety of causal questions from purely observational data.
For example, informative features used by a model, informative features not used by a model, suppressors, and non-suppressor non-informative features can be distinguished by combining multivariate ML modelling with univariate statistical tests \citep{weichwald2015causal}.
This forms the basis of the Pattern \citep{haufeInterpretationWeightVectors2014} and PatternGAM \citep{clarkcorrecting} approaches.

In the balanced settings of both studies presented, manipulated image parts that are not in themselves informative are possible suppressors.
Given the tendency of feature attribution methods to highlight both informative and suppressor features \citep{wilmingTheoreticalBehavior2023}, elevated RIW and RIL could be expected in both the confounded and balanced settings, as opposed to the no-watermark and baseline settings under hypothesis H2.
Instead, no substantial differences in RIW and RIL between the balanced and the other two settings are observed.
This suggests that for the prediction tasks studied, the causal structure of the training data, the resulting observable statistical properties of the data, and ultimately the model function itself, play only a minor role when it comes to attributing importance, compared to low-level properties of individual test images such as salience.

\subsection*{Explainable AI and distribution shifts}
Shortcut learning -- that is, training models on statistical associations induced by confounding -- is not inherently problematic.
However, unstable causal relationships may induce distribution shifts that degrade generalisation in testing (e.g., production) systems.
In both studies, we have demonstrated poor generalisation if a model trained in the confounded setting is applied to data for which either the prevalence $P(C)$ of the confounder differs (e.g., when no pictures taken by the photographer in question are included in the test set), or where the causal influence $P(Y|C)$ of the photographer on the image category differs (e.g., the photographer's preference changes from taking pictures of dogs to taking equal amounts of pictures of cats and dogs).
Similarly, we expect that changes in the causal influence $P(W | C)$ of the photographer on the confounded input feature (e.g., if the photographer changes the appearance of their watermark) would also lead to decreases in model performance.

Feature attribution methods, typically applied in the context of a single trained model, are not intended to detect distribution shifts.
Using XAI to detect shortcuts requires the user to determine not only if a feature is a confounder, but whether that confounding relationship is stable from training to possible production regimes.
Confounding effects can be highly stable if both causal links $C \rightarrow X_i$ and $C \rightarrow Y$ represent natural phenomena that can be considered invariant across across domain shifts.
In fact, machine learning problems are frequently set up to rely on confounded statistical associations. 
An example would be biomarker discovery using machine learning \citep[e.g.,][]{tideman_automated_2021}.
Here, a genetic disposition might cause behavioural symptoms that give rise to a diagnosis on one hand, and aberrations in medical images of an affected patient on the other hand.
In a hypothetical setting where genetic sequencing may be considered too costly, one may be interested in deriving an imaging-based biomarker for the disease, by training a machine learning model to predict the diagnostic label.
As both causal links establishing the confounding relation in this case represent stable biological mechanism, one might expect such a model to be transferable to, for example, different demographic cohorts.

Conversely, not only distribution changes due to unstable confounding, but \emph{any} changes in distributions involving either the target itself or variables used by a model (for predicting the target) can endanger a model's generalisation ability.
This would not only be the case for variables affected by confounding but also for variables causally or anti-causally linked to the target, as well as for suppressor variables.
The impact of different types of distribution shifts as well as of possible mitigation measures are studied in fields such as domain adaptation and transfer learning \citep{pan2010Transfer,zhuang2021Transfer}. 
To study generalisation, a causal perspective can again be useful. Knowledge of the causal graph linking features $X$ and target $Y$ provides a natural factorisation of their joint distribution $P(X,Y)$ into a product of lower-dimensional marginal and conditional distributions, each of which corresponds to an independent causal mechanism \citep{scholkopfCausalityMachineLearning2019,castro2020causality}.
With this, distribution shifts, their influence on machine learning models, and the impact of possible mitigation measures can be investigated separately for each causal mechanism through theory and controlled experimentation, while ensuring that the causal structure of the data is not violated. This has been demonstrated here through manipulating $P(Y|C)$ and $P(C)$.

\subsection*{Implications for downstream XAI goals}
\label{sec:downstream}
There is ample theoretical and empirical evidence showing that popular attribution methods assign importance to non-informative suppressors \citep{haufeInterpretationWeightVectors2014,kindermansLearningHowExplain2017,wilmingScrutinizingXAIUsing2022,wilmingTheoreticalBehavior2023}, severely diminishing their utility for model diagnostics, scientific discovery, and algorithmic recourse \citep{haufe2024position}.
Consequently, a watermark acting purely as a suppressor without being in itself task-informative could still be attributed significant importance due to being useful for the model. 
Such attribution would not be equivalent to shortcut learning, but the watermark would rather constitute a nuisance that may degrade model performance.
A trained model may thus attempt to remove the watermark by subtracting informative and non-informative pixels covered by the watermark.
Notably, a model trained to suppress undesired variance does not necessarily suffer a performance drop when applied to data lacking the suppressor.
As evidenced in Tables~\ref{tab:AUROCs}, \ref{tab:AUROCs-coco}, \ref{tab:AUROCs-variable}, and \ref{tab:AUROCs-inverted}, models trained in the balanced setting perform equally well when applied to either balanced data, confounded data, or data without watermark or lightness manipulations applied. 
Thus, dismissing or attempting to `correct' a model that uses watermark structures for suppression in the balanced setting would be neither necessary nor advisable and might result in an avoidable drop in performance.
Feature attributions methods alone cannot provide the necessary guidance to distinguish this case from actual shortcut learning.

Similar considerations can be made for models using protected attributes for suppression.
A model including race or sex for the sole purpose of removing the influence of these factors on predictions of recidivism risk may be generally desirable, not only legally and ethically, as discussed in \citep{clarkcorrecting}.
Finally, when the goal is to identify features causally driving the target to enable scientific discoveries \citep{samek2019towardsexplainable, jimenezLunaDrugDiscoveryExplainable2020,tideman_automated_2021,watson_interpretable_2022,wong_discovery_2024}, or to identify causal interventions changing the target for algorithmic recourse \citep{ustun_actionable_2019, karimi2021algorithmic}, then the use of suppressor-attributing feature attribution methods will systematically lead to false-positive detections and causally infeasible interventions \citep{haufe2024position}.

Going beyond these considerations, the present study demonstrates that not only attribution to suppressors poses a threat to the interpretation of feature attribution methods. The observed attributions show a far stronger dependence on low-level properties of test inputs, which are entirely independent of the model under consideration by construction. Based on this result, the utility of the investigated methods to provide quality assurance for machine learning models must be further questioned. Likewise, the validity of complex quality assurance workflows using feature attribution as a building block, e.g. for confounder removal, may need to be critically re-evaluated.

\subsection*{Structural challenges in XAI research}
This study reports a dominant dependency of feature attributions on structural image properties that are unrelated to the statistical associations between features and target learned and utilised by machine learning models. Previous research has further demonstrated that feature attribution methods i) are bound to systematically assign importance to non-informative suppressor features (see \nameref{sec:downstream} and the references therein), ii) are unstable across parameter choices and inconsistent with each other \citep{babic2021beware,hedstrom2023meta,bluecher2024decoupling}, and iii), are vulnerable to adversarial manipulation \citep{dombrowskiExplanationsCanBe2019}.

\citet{haufe2024position} argue that these failures stem from a lack of formal problem definitions aligned with explanation goals like the detection of shortcut learning, precluding theoretical analyses of methods with respect to these goals. 
Consequently, the field is also lacking formal notions of explanation correctness that can be operationalised and empirically evaluated.
XAI methods are frequently defined purely algorithmically based on informal intuitions that may be incorrect. An example is the widespread misunderstanding that multivariate models use only informative features for prediction, which holds only in simplistic settings where all features either causally drive the target or are entirely causally separated from it. 

Validition is typically demonstrated in two steps.
First, an XAI method's `faithfulness' is quantified by measuring the drop in model performance induced by omitting or randomising features with high importance \citep[e.g.,][]{jacoviFaithfullyInterpretableNLP2020}, where methods experiencing larger drops are considered better.
\citet{wilmingTheoreticalBehavior2023} have shown, though, that faithfulness metrics incentivise attribution to suppressors; thus, they do not represent viable metrics of explanation performance for purposes such as shortcut learning detection. 
Second, the utility of the method for a desired purpose, such as shortcut learning detection, is demonstrated qualitatively for a range of examples. 
This practice can also lead to a biased assessment of the method's true capabilities, though. 
On one hand, as such demonstrations often remain purely qualitative, the choice of the data samples used for visual inspection and reporting may strongly influence the conclusions drawn. 
On the other hand, such demonstrations (qualitative {or} quantitative), could be confounded by the choice of the prediction problems analysed.
Prediction problems, for which informative features coincide with features susceptible to strong attribution through their structural properties, could then lead to overly optimistic assessments. 
This is what appears to be the case in the two image classification problems studied here. 

To counteract biased assessments of XAI aptitude for specific explanation goals, \citet{haufe2024position} advocate a research process by which XAI problems are formally defined in alignment with desired explanation goals \citep[see also][]{DIN_SPEC_92001-3:2023-04}, and where methods are designed and evaluated accordingly.
We argue that the explanation goals discussed here inherently require knowledge about the causal roles of target and individual features within the data-generating process -- information that cannot be provided by current feature attribution methods.

\subsection*{Limitations and Outlook}

In this study, we controlled only specific features (watermarks or lightness) but not all information contained in the studied images. In that, we did not control the task-related information contained in other image features, notably the objects to be classified themselves. We refrained from a completely controlled setting, because, first, complete knowledge about all task-related features is not available in real data \citep[see][]{haufe2024position}; second, studies controlling all task-related information have already experimentally demonstrated the widespread tendency of XAI methods to attribute importance to features not part of the ground truth \citep{wilmingScrutinizingXAIUsing2022,clark2024tetris,oliveiraPretraining2024,wilming2025gecobench}; and, third, controlling watermarks as a practically relevant artifact of real data allowed us to relate our systematic findings directly to observations made in \citet{lapuschkinUnmaskingCleverHans2019}. In principle, though, we cannot exclude that task-informative objects (cats, dogs, cars, or animals) receive similarly high or even higher attribution for certain images. Future studies could parametrically study the impact of feature salience relative to other properties such as task-informativeness, suppression, and out-of-distribution characteristics. The high relative importance attributed to salient watermarks and lightness features in all studied settings (often two to three times higher on average than the mean attribution across the entire image), however, strongly suggests that those features would be considered relevant by practitioners inspecting the generated attribution maps in any case, inviting various misinterpretations.

As the focus of this work is on benchmarking feature attribution methods, we refrained from testing each of these for a large battery of different underlying machine learning model architectures. Rather, we focused on a standard convolutional neural networks (CNN) architecture that is widely used in image classification and yielded high classification accuracy for all of the studied task. In doing so, our results are comparable to that of other XAI studies reporting on highly similar architectures \citep{lapuschkinUnmaskingCleverHans2019,andersRemovingHans2022}. Our results obtained for CNN models strongly challenge the general claim that feature attribution methods can systematically reveal what machines have learned. Future studies may be able to provide a more nuanced assessment by differentiating between model architectures. However, in absence of a theory explaining the behaviour of feature attribution methods as a function of structural properties of the data-generating process and the machine learning model, observing improvements for certain settings would still offer little guidance for users of XAI methods.

Note also that we used standard implementations of all feature attribution methods with
all parameters set to the packages' default values, thereby often following direct recommendations by the original methods' developers. 
We deliberately did not perform parameter sweeps or optimisation as proposed in some recent work \citep[e.g.,][]{pahde2023optimizing}. 
This allowed for consistency with \citet{lapuschkinUnmaskingCleverHans2019} who also investigated importance attribution to watermarked images.
However, optimising for the detection of class-related structures using ground-truth annotations, as proposed in \citet{pahde2023optimizing}, is typically not possible in real-world applications of XAI \citep[see also the general criticism of human-annotated ground-truth data for XAI validation in][]{haufe2024position}.
Also, introducing preferences for certain XAI outputs (such as depictions of cats) precludes several key explanation purposes such as scientific discovery \citep{samek2019towardsexplainable, jimenezLunaDrugDiscoveryExplainable2020,tideman_automated_2021,watson_interpretable_2022,wong_discovery_2024} and model or data diagnostics, which require these outputs to be independent of ground-truth information.
As an example, an attribution method optimised to highlight pre-annotated lesions in histopathology might only serve as confirmatory evidence, but is biased when it comes to judging a model's fitness-for-purpose.
Such a method may also be incapable of identifying task-related features not known to the annotator (a type II error arising from the labelling process), even if the model uses these.

\section{Methods}

\subsection{Data generation -- watermark study}
\label{sec:data_watermarks}
For the watermark study, we sourced $N=9,600$ images of cats and dogs ($N_{\text{cat}} = N_{\text{dog}} = 4,800$) from Kaggle \citep{kaggle2018catsdogs}. Images were stored and processed in RGB colour space. All images were resized to $128 \times 128$-pixels. Each flattened sample image, $\mathbf{x} = [{\mathbf{x}^{\text{R}}}^\top, {\mathbf{x}^{\text{G}}}^\top, {\mathbf{x}^{\text{B}}}^\top]^\top \in \mathbb{R}^{3 \cdot D}$ with $D = 128^2$, was rescaled to hold values between 0 and 1 in each of the three colour channels using the Min-Max scaler
\begin{equation}
    \mathbf{x}^{\text{R/G/B}} = \frac{\mathbf{x}^{\text{R/G/B}} - \min (\mathbf{x}^{\text{R/G/B}})}{\max (\mathbf{x}^{\text{R/G/B}}) - \min (\mathbf{x}^{\text{R/G/B}}) } \;,
    \label{eq:watermarks-scaling}
\end{equation}
where 0 encodes darkest and 1 encodes lightest regions. 
We refer to the resulting dataset as the \emph{no-watermark} or baseline dataset $\mathcal{D}_{\text{no-wm}}$. Subsequently, we applied a translucent watermark to varying degrees of prevalence to create two more datasets: a \emph{confounded} dataset $\mathcal{D}_{\text{conf}}$, where 80\% of randomly-selected dog images but only 20\% of cat images contain the watermark, and a \emph{balanced} dataset $\mathcal{D}_{\text{bal}}$, in which 50\% of the images of each class contain the watermark. 

The watermark consists of a Q-shaped logo in greyscale with accompanying text `Quality in Artificial Intelligence, QAI LABS' within a 128 $\times$ 128 pixel mask. Logo and text take greyscale values between 0.2 and 0.4, while the background is white (greyscale 1). The total number of pixels in the watermark foreground is $|\mathcal{W}| = 1983$. 
The watermark was added either in a fixed or a variable location for each sample to be affected. In the fixed-position case, it was placed centrally in the top part of the image. In the variable-location case, the location was randomly selected for each sample out of all feasible locations ensuring that the entire watermark mask is strictly contained in the image. 
In both cases, the watermark was combined with the existing image by multiplication of the greyscale watermark with the inverted original sample image, after which the combined image was inverted again. 
With this, we obtain a salient light watermark with 80\% maximum opacity in each pixel of the affected image.

Using five different fixed random seeds, we created five splits of each dataset, each containing different watermarked and non-watermarked samples of both classes. 
Each split yielded $6,720$ training samples  $\mathcal{D}^{\text{train}}$ and $1,440$ samples each for the validation and test sets $\mathcal{D}^{\text{val}}$ and $\mathcal{D}^{\text{test}}$, respectively, amounting to 70/15/15 percent splits of the overall data.
All datasets contain a balanced amount of cat and dog samples, and (within a given split) the underlying images are identical across the three datasets. For example, a cat image in the confounded training set will not be in any of the confounded, balanced, or no-watermark test sets. 

We also created \emph{all-watermark} test sets $\mathcal{D}^{\text{test}}_{\text{wm}}$ by applying the watermark to each and every image of the no-watermark set.
This allows us to compare attributions for a given model on identical test images that only differ in the presence or absence of the watermark.

Finally, we also created variants of all datasets with \emph{inverted colour encoding}, where 1 encodes darkest and 0 encodes lightest regions. 

\subsection{Data generation -- lightness study}
\label{sec:data_lightness}

In the lightness study, we make use of an animal versus vehicle `supercategory' binary image classification problem utilising the Common Objects in Context \citep[COCO,][]{tsung2014microsoftcoco} dataset. Here, we sourced $N=30,000$ ($N_{\text{animal}} = N_{\text{vehicle}} = 15,000$) images from COCO, with each sample re-scaled to be $224 \times 224$-pixels in size; thus the overall image dimensionality is $D = 224^2$.
COCO provides category and supercategory labels, which we used to ensure that no sample contains both an animal and a vehicle and thus belongs to more than one of the classes.

Each sample was converted from the original RGB to HLS (Hue, Lightness, Saturation) colour space using the OpenCV package in Python for manipulation of the lightness channel and subsequent analysis. 
Hue is a circular variable which is encoded into the range $[0,180]$ by OpenCV, for degrees 0 to 360 in steps of 2, whereas Lightness and Saturation are in the range $[0,255]$. 
As such, we divided the Hue channel values by 180 and Lightness and Saturation by 255 to scale all channels to the range $[0,1]$. 
We proceeded to manipulate the lightness channel using histogram matching \citep{borke2011histogram} of the pixel values of the lightness channel with a reference histogram sampled from three differently parametrised Beta distributions $\text{Beta}(\alpha, \beta)$. 
Exemplary results of this histogram matching process can be seen in Figure~\ref{fig:data-coco}.

We created a \emph{baseline} dataset $\mathcal{D}_{\text{base}}$ in which the lightness channel of all images of both classes was normalised by transforming the lightness value distribution into the central $\text{Beta}(3,3)$ distribution. This dataset is conceptually analogous to the {no-watermark} dataset in the watermark study.
Again in analogy to the watermark study, we created two additional datasets $\mathcal{D}_{\text{conf}}$ and $\mathcal{D}_{\text{bal}}$ by manipulating the lightness channel. For the \emph{confounded} dataset $\mathcal{D}_{\text{conf}}$, we darkened 50\% of the samples of the animal class by transforming their lightness histograms into the left-skewed $\text{Beta}(2,4)$ distribution, and we lightened 50\% of the samples of the vehicle class by transforming their histograms into the right-skewed $\text{Beta}(4,2)$ distribution.
For the balanced dataset $\mathcal{D}_{\text{bal}}$, 25\% of the samples of both classes were darkened, 25\% were lightened, and 50\% were normalised by enforcing respective $\text{Beta}(2,4)$, $\text{Beta}(4,2)$, and $\text{Beta}(3,3)$ lightness distributions. 
All remaining samples of the confounded and balanced datasets were normalised to the $\text{Beta}(3,3)$ distribution. 

We again created five `shufflings' of the datasets through the use of different fixed random seeds.
Each resulting training split  $\mathcal{D}^{\text{train}}$ thereby contains 21,000 samples, whereas corresponding validation and test splits $\mathcal{D}^{\text{val}}$ and $\mathcal{D}^{\text{test}}$ contain 4,500 samples each, again according to a 70/15/15 percent split.

Finally, again in analogy to the watermark study, we also created variants $\mathcal{D}^{\text{test}}_{\text{dark}}$ and $\mathcal{D}^{\text{test}}_{\text{bright}}$ of the test splits, in which the respective lightness manipulations were applied to all samples.
\subsection{Model training and evaluation}
\label{sec:training}

With the datasets defined for both studies as well as for two different watermark positioning schemes and two different colour encodings in the watermark study, we trained five ML models $f^{\boldsymbol{\theta}}:\mathbb{R}^D \rightarrow \mathcal{Y}$ parameterised by vectors $\boldsymbol{\theta}$ with different initialisations as per differing fixed random seeds. This was done for each data shuffling using the training ($\mathcal{D}^{\text{train}})$ and validation ($\mathcal{D}^{\text{val}})$) splits of the dataset $\mathcal{D}_{\text{no-wm}}$, $\mathcal{D}_{\text{conf}}$, and $\mathcal{D}_{\text{bal}}$ for the watermark study and  of the datasets $\mathcal{D}_{\text{base}}$, $\mathcal{D}_{\text{conf}}$, and $\mathcal{D}_{\text{bal}}$ for the lightness study.

A classical ReLU-activated convolutional neural network (CNN) architecture was used, implemented in PyTorch version 2.6.0, with three rounds of convolutions followed by max pooling, ending with two dropout layers (just one in the case of the lightness models) and two fully connected layers leading to the final output layer with one neuron per class. 
The first convolution layer takes in the three-channel (representing RGB values in the watermark study and HLS values in the lightness study) $D$-dimensional images and outputs 64 channels, with each convolution producing progressively higher channel counts [3, 64, 128, 256], with kernel sizes of 3 in each layer for the watermark models, and [5,3,5] respectively for the lightness models. 
Batch normalisation was applied after each convolution for the watermarks models.
The three max pooling operations all use kernel sizes of 2 with strides of 2 as well for the watermarks models, and kernel sizes of [2,4,4] respectively for the lightness models.

We used stochastic gradient descent (SGD) optimisation with a learning rate of $0.005$, momentum of $0.9$, a batch size of 64, and a weight decay of $0.001$ and $5 \times 10^{-5}$ for the watermark  and lightness experiments, respectively.
Models were trained on training data $\mathcal{D}_{\text{train}}$ over $100$ epochs with cross-entropy loss, and performance was evaluated after each epoch on separate validation data $\mathcal{D}^{\text{val}}$ using the area under the receiver-operating characteristic curve (AUROC) metric.
Model with highest validation performance over all epochs were selected as final models. Final models trained in the confounded, balanced and no-watermark/baseline settings were evaluated on the test splits $\mathcal{D}^{\text{test}}_{\text{no-wm}}$, $\mathcal{D}^{\text{test}}_{\text{conf}}$, $\mathcal{D}^{\text{test}}_{\text{bal}}$, and $\mathcal{D}^{\text{test}}_{\text{wm}}$ for the watermark study and $\mathcal{D}^{\text{test}}_{\text{base}}$, $\mathcal{D}^{\text{test}}_{\text{conf}}$, $\mathcal{D}^{\text{test}}_{\text{bal}}$, $\mathcal{D}^{\text{test}}_{\text{dark}}$, and $\mathcal{D}^{\text{test}}_{\text{bright}}$ for the lightness study. The resulting AUROC scores were averaged across test splits.

%
\subsection{Feature attribution}
\label{sec:attribution}
The feature attribution methods explored in this work are Deconvolution \citep{zeilerVisualizingUnderstandingConvolutional2014}, Integrated Gradients \citep{sundararajanAxiomaticAttributionDeep2017}, Gradient SHAP \citep{lundbergUnifiedApproachInterpreting2017}, and Layer-wise Relevance Propagation (LRP) using the LRP-$\epsilon$ and LRP-$\alpha \beta$ variants \citep{bachPixelWiseExplanationsNonLinear2015}. 
For the first four methods, we used implementations from the Captum library version 0.7.0 \citep{kokhlikyanCaptumUnifiedGeneric2020} with their default values and the zero-input baseline for SHAP. 
LRP-$\alpha \beta$ \citep{bachPixelWiseExplanationsNonLinear2015} was implemented using the Zennit framework\footnote{https://github.com/chr5tphr/zennit} with the default settings $\alpha=2, \beta=-1$. Identical values were used in \citep{lapuschkinUnmaskingCleverHans2019} for CNN-based classification of ImageNet \citep{deng2009imagenet} data.

A model $f^{\boldsymbol{\theta}}:\mathbb{R}^D \rightarrow \mathbb{R}$, trained according to parameterisation $\boldsymbol{\theta}$ over $\mathcal{D}^{\text{train}}$, was provided as input to each feature attribution method. Also provided was the test sample $\mathbf{x} \in \mathcal{D}^{\text{test}}$ to be `explained' as well as the zero-input $\mathbf{b} = \boldsymbol{0}$ serving as a reference point for SHAP. 

Each method produces a feature attribution map $\mathbf{a}(f^{\boldsymbol{\theta}}, \mathbf{x}_{\text{test}}^{(n)}, \mathbf{b}) \in \mathbb{R}^{3 \cdot D}$, reduced subsequently to just $\mathbf{a}$ for ease of reference. 
We further refer to individual RGB channels in the watermark study and individual HLS channels in the lightness study as $\mathbf{a}^{\text{R/G/B}} \in \mathbb{R}^D$ and $\mathbf{a}^{\text{H/L/S}} \in \mathbb{R}^D$, respectively.

In addition to the five model-dependent feature attribution methods, we also considered two model-independent baselines. First, the Laplace edge-detection filter kernel $\mathcal{S}$ is defined as
\begin{equation}
    \mathcal{S} = \begin{bmatrix} 0 & 1 & 0 \\ 1 & -4 & 1 \\ 0 & 1 & 0 \end{bmatrix}
\end{equation}
and was applied separately to each colour channel of the input image via 2D convolution: $\mathbf{a}_{\text{Laplace}}^{\text{R/G/B}} = \mathcal{S} \ast \mathbf{x}^{\text{R/G/B}}$. Second, we also considered the raw input images $\mathbf{x}$ themselves as their own explanation for some analyses.

We finally define the pixel-wise average importance over all colour channels as $\mathbf{a}^{\text{RGB}} = \nicefrac{1}{3}(|\mathbf{a}^{\text{R}}| + | \mathbf{a}^{\text{G}} | + | \mathbf{a}^{\text{B}} |) \in \mathbb{R}^D$ and $\mathbf{a}^{\text{HLS}} = \nicefrac{1}{3}(|\mathbf{a}^{\text{H}}| + | \mathbf{a}^{\text{L}} | + | \mathbf{a}^{\text{S}} |) \in \mathbb{R}^D$, respectively, where $|\cdot|$ denotes taking the absolute value element-wise.

For each of the five models, for each data shuffling, and for each training setting (confounded, balanced, and no-watermark/baseline), feature attributions were obtained. 
This was done using each of the feature attribution methods for each test sample in the $\mathcal{D}^{\text{test}}_{\text{no-wm}}$ and $\mathcal{D}^{\text{test}}_{\text{wm}}$ datasets for the watermark study and in the $\mathcal{D}^{\text{test}}_{\text{base}}$, $\mathcal{D}^{\text{test}}_{\text{dark}}$, and $\mathcal{D}^{\text{test}}_{\text{bright}}$ datasets for the lightness study.
. 

\subsection{Quantification of attribution to watermarks and pixel lightness}
We quantified the normalised proportion of importance attributed to the image parts targeted by experimental manipulations (the watermarked area in the watermark study and the lightness channel in the lightness study). This was done for each test image, regardless of whether a manipulation was actually carried out for that image.
Recall that, depending on the studied experimental setting, manipulated image parts could represent either informative or non-informative (suppressor) variables with respect to the model's prediction target, or outliers with respect to the training distribution. 

For a given attribution $\mathbf{a} \in \mathbb{R}^D$ in the watermark study, we calculated the sum of the absolute importances in all three colour channels attributed to pixels within the watermarked area ($\mathcal{W}$) over the total sum of absolutes importances within the entire image. 
These values were divided by the respective numbers of pixels in the watermark ($D_{w}$) and in the full image ($D = 128^2$), such that the resulting \emph{relative importance on watermark} 
\begin{equation}
    \text{RIW} = \frac{\nicefrac{1}{D_w}\sum_{w \in \mathcal{W}}a_w^{\text{RGB}}}{\nicefrac{1}{D}\sum_{i \in \{1, \hdots, D\}}a_i^{\text{RGB}}}
\end{equation}
is normalised by the size of the watermark. With this, RIW quantifies how much more importance, on average, is assigned to pixels belonging to the watermark compared to the average per-pixel importance over the entire image. Hence a value of RIW $=2$ would mean that the pixels belonging to the watermark are, on average, attributed twice as much importance as the average pixel in the entire image.

In the lightness study, we analogously evaluated the proportion of importance attributed globally to the lightness channel compared to the average importance attributed globally to all three HLS channels together. This \emph{relative importance on lightness} is defined as
\begin{equation}
    \text{RIL} = \frac{\sum_{i \in \{1, \hdots, D\}}|a_i^{\text{L}}|}{\sum_{i \in \{1, \hdots, D\}}a_i^{\text{HLS}}} \;,
\end{equation}
where the sum of the absolute-valued attribution to the lightness channel is divided by the sum of the mean absolute attribution value across all channels  across the whole image. Analogous to RIW, a value of RIL $=2$ would indicate that, across the entire image, lightness features are attributed twice as much importance as the average of the three HLS channels.



\subsection{Singular value decomposition analysis}
Complementary to the RIW metric, we applied singular value decomposition (SVD) to investigate to what extent watermarks in test images dominate attribution maps in the confounded, balanced, and no-watermark training settings of the watermark study. This analysis was carried out for all considered attribution methods for fixed-position watermarks only. For each training setting and attribution method, attributions were obtained for all $N^{\text{test}} = 1,440$ test images with watermarks applied for a single data split and model fit per training setting.
We vectorised each RGB attribution by concatenating the three channels into the feature axis, yielding $\mathbf{a}(n) \in \mathbb{R}^{3 \cdot D}$, after which the mean across samples was subtracted: $\mathbf{a}(n) \leftarrow \mathbf{a}(n) - \bm{\mu}_\mathbf{a}$. 
Attributions for all samples were concatenated to form a matrix $\mathbf{A} = [\mathbf{a}(1), \hdots, \mathbf{a}(N_{\text{test}})] \in \mathbb{R}^{3 \cdot D \times N^{\text{test}}}$, 
which was decomposed as $\mathbf{A} = \mathbf{U S V}^\top$, where $\mathbf{U}$ is a matrix of singular vectors, $\mathbf{V}$ is a matrix of corresponding factors, and $\mathbf{S}$ is a diagonal matrix containing the corresponding singular values in decreasing order \citep{golub2013matrix}. The first singular vector $\mathbf{u}_1 = [{\mathbf{u}_1^{\text{R}}}^\top, {\mathbf{u}_1^{\text{G}}}^\top, {\mathbf{u}_1^{\text{B}}}^\top]^\top \in \mathbb{R}^{3 \cdot D}$, corresponding to the largest singular value, represents the spatial profile of the component accounting for the largest variance in the image across all samples. 
We calculated the average absolute-valued spatial contribution across the three colour channels as $\mathbf{a}_{\mathbf{u}_1}^{\text{RGB}} = \nicefrac{1}{3}(|\mathbf{u}_1^{\text{R}}| + | \mathbf{u}_1^{\text{G}} | + | \mathbf{u}_1^{\text{B}} |) \in \mathbb{R}^D$, and reshaped the resulting vector into a 128 $\times$ 128 pixel matrix for visualisation. 

For comparison, we also applied SVD in the same manner to the 
corresponding raw data matrix $\mathbf{X} = [\mathbf{x}(1), \hdots, \mathbf{x}(N^{\text{test}})] \in \mathbb{R}^{3 \cdot D \times N^{\text{test}}}$ of the watermarked test images for which the feature attributions have been derived. We again concatenated RGB channels into the feature axis, removed the column-wise means,
and performed SVD. This resulted in analogous images $\mathbf{x}_{\mathbf{u}_1}^{\text{RGB}} \in \mathbb{R}^D$ to be visualised for comparison.

\subsection{Statistical analyses}

Histograms of the RIW distributions for test images with and without watermark ($\mathcal{D}^{\text{test}}_{\text{wm}}$ and $\mathcal{D}^{\text{test}}_{\text{no-wm}}$ datasets) in the watermark study are shown in Figure~\ref{fig:hist_watermark_fixed} for the fixed-position case, in Supplemental Figure~\ref{fig:hist_watermark_variable} for the variable-position case, and in Supplemental Figure~\ref{fig:hist_watermark_inverted} for the inverted colour encoding case.
Analogous histograms of RIL distributions for test images with and without lightness manipulations applied ($\mathcal{D}^{\text{test}}_{\text{base}}$, $\mathcal{D}^{\text{test}}_{\text{dark}}$, and $\mathcal{D}^{\text{test}}_{\text{bright}}$ datasets) are shown in Figure~\ref{fig:hist_lightness} for the lightness study. 

Further statistical assessment was carried out for the watermark study. For this, RIW scores resulting from a single data split were averaged across the five ML model fits, resulting in $N^{\text{test}} = 1,440$ scores each for watermarked and non-watermarked test images, respectively, for each training setting and attribution method, half of which were obtained from cat and dog images, respectively. Means and standard errors of the RIW metric are shown in Supplementary Table~\ref{tab:mean_se} for all combinations of training settings, test image characteristics (image class, watermark presence), and feature attribution methods.
The effects of training setting and test image characteristics on RIW was further assessed using linear mixed models (LMM) as implemented by Matlab's fitlme function. Since, for a given test image, setting (confounded, balanced, no-watermark, one-hot encoded using two binary regressors) and watermark (present or not) are orthogonal within-factors by design, separate models (RIW $\sim$ setting and RIW $\sim$ WM in LMM notation) could be fitted to assess the amount of variation in RIW (measured in terms of $R^2$) explained by either setting or watermark presence. Models were fitted on the pooled data of all three settings but also separately on all pairwise combinations of two settings, which allowed for a simple way to assess the direction and statistical significance of any differences in RIW between settings.
Note that image class itself was not investigated as a separate factor since significant RIW differences between dogs and cats were observed for multiple attribution methods including the model-independent Laplace filter in various settings, speaking to differences in the image statistics of both classes that were not controlled for.

\section*{Data and Code availability}
The datasets used in the watermarks \citep{kaggle2018catsdogs} and lightness \citep{tsung2014microsoftcoco} studies are publicly available for download.


All source code used to generate data and conduct the experiments in both studies is available (\url{https://github.com/braindatalab/debugging_xai}).
Where possible, we have used fixed seeding to ensure reproducibility of generated data and experimental results.
Exact data and results files can be provided on request.

\section*{Funding}
This result is part of a project that has received funding from the European Research Council (ERC) under the European Union’s Horizon 2020 research and innovation programme (Grant No. 758985), and the German Federal Ministry for Economic Affairs and Climate Action (BMWK) within the ``Metrology for Artificial Intelligence in Medicine (M4AIM)'' program of the ``QI-Digital'' initiative.

\bibliography{xai_better}
\bibliographystyle{icml2023}

\newpage

\begin{center}
\begin{doublespace}
{\Large Feature salience -- not task-informativeness -- \\drives machine learning model explanations}\\
\vspace{12pt}
{\Large Supplementary Material}
\end{doublespace}
\vspace{12pt}
{\large Benedict Clark, Marta Oliveira, Rick Wilming, Stefan Haufe}\\
\vspace{6pt}
\date{February 2023}
\end{center}

\renewcommand{\thesection}{S\arabic{section}}
\renewcommand{\thetable}{S\arabic{table}}
\renewcommand{\thefigure}{S\arabic{figure}}

\setcounter{section}{0}
\setcounter{table}{0}
\setcounter{figure}{0}


\begin{table}[htbp]
\sffamily
\begin{small}
\tabcolsep1.5mm
\begin{tabular}{lccrrrrrp{1mm}rr}
    \toprule
    Model & \multicolumn{2}{c}{Test Image} & \multicolumn{8}{c}{RIW (\textbf{Mean}$\pm$SE)}\\
    & WM & Cat & \multicolumn{1}{c}{Deconv} & \multicolumn{1}{c}{IG} & \multicolumn{1}{c}{GradSHAP} & \multicolumn{1}{c}{LRP-\(\epsilon\)} & \multicolumn{1}{c}{LRP-\(\alpha\beta\)} & & \multicolumn{1}{c}{Laplace} & \multicolumn{1}{c}{\textbf{x}} \rule{0pt}{3ex}\\
    \midrule
    \multicolumn{1}{c}{\multirow{6}{*}{\rotatebox{90}{Confounded}}} & \checkmark & \checkmark & \textbf{1.38}$\pm$0.004 & \textbf{1.89}$\pm$0.009 & \textbf{1.87}$\pm$0.010 & \textbf{1.97}$\pm$0.009 & \textbf{2.11}$\pm$0.012 & \multirow{18}{*}{\resizebox{2.5mm}{4.25cm}{$\Biggr\}$}} & &\\
     & \checkmark &  & \textbf{1.36}$\pm$0.004 & \textbf{1.80}$\pm$0.008 & \textbf{1.79}$\pm$0.009 & \textbf{1.84}$\pm$0.008 & \textbf{1.94}$\pm$0.011 &  & & \\
     &  & \checkmark & \textbf{1.06}$\pm$0.002 & \textbf{1.11}$\pm$0.007 & \textbf{1.10}$\pm$0.008 & \textbf{1.17}$\pm$0.008 & \textbf{1.01}$\pm$0.007 & & &\\
     &  &  & \textbf{1.04}$\pm$0.002 & \textbf{1.11}$\pm$0.007 & \textbf{1.11}$\pm$0.007 & \textbf{1.15}$\pm$0.007 & \textbf{1.00}$\pm$0.007 & & & \\
     & $\Delta$ & \checkmark & \textbf{0.33}$\pm$0.004 & \textbf{0.78}$\pm$0.010 & \textbf{0.76}$\pm$0.011 & \textbf{0.80}$\pm$0.011 & \textbf{1.10}$\pm$0.012 &  & & \\
     & $\Delta$ &  & \textbf{0.32}$\pm$0.004 & \textbf{0.69}$\pm$0.009 & \textbf{0.68}$\pm$0.009 & \textbf{0.69}$\pm$0.009 & \textbf{0.93}$\pm$0.010 &  & & \\
     \cmidrule{1-8}
    \multicolumn{1}{c}{\multirow{6}{*}{\rotatebox{90}{Balanced}}} & \checkmark & \checkmark & \textbf{1.36}$\pm$0.005 & \textbf{1.72}$\pm$0.009 & \textbf{1.70}$\pm$0.009 & \textbf{1.76}$\pm$0.009 & \textbf{1.77}$\pm$0.014 &  & \textbf{1.79}$\pm$0.019 & \textbf{1.40}$\pm$0.008\\
     & \checkmark &  & \textbf{1.21}$\pm$0.005 & \textbf{1.64}$\pm$0.008 & \textbf{1.63}$\pm$0.008 & \textbf{1.65}$\pm$0.008 & \textbf{1.62}$\pm$0.012 & & \textbf{1.58}$\pm$0.018 & \textbf{1.39}$\pm$0.007\\
     &  & \checkmark & \textbf{1.08}$\pm$0.003 & \textbf{1.03}$\pm$0.007 & \textbf{1.03}$\pm$0.007 & \textbf{1.08}$\pm$0.007 & \textbf{0.99}$\pm$0.007 & & \textbf{1.07}$\pm$0.010 & \textbf{0.97}$\pm$0.008\\
     &  &  & \textbf{1.01}$\pm$0.003 & \textbf{1.04}$\pm$0.007 & \textbf{1.04}$\pm$0.007 & \textbf{1.06}$\pm$0.006 & \textbf{0.97}$\pm$0.007 & & \textbf{1.05}$\pm$0.008 & \textbf{0.98}$\pm$0.007\\
     & $\Delta$ & \checkmark & \textbf{0.28}$\pm$0.004 & \textbf{0.69}$\pm$0.010 & \textbf{0.67}$\pm$0.010 & \textbf{0.68}$\pm$0.010 & \textbf{0.79}$\pm$0.013 & &\textbf{0.72}$\pm$0.019 & \textbf{0.42}$\pm$0.008\\
     & $\Delta$ &  & \textbf{0.20}$\pm$0.004 & \textbf{0.60}$\pm$0.008 & \textbf{0.59}$\pm$0.008 & \textbf{0.59}$\pm$0.009 & \textbf{0.65}$\pm$0.012 & & \textbf{0.53}$\pm$0.017 & \textbf{0.41}$\pm$0.006\\
     \cmidrule{1-8}
    \multicolumn{1}{c}{\multirow{6}{*}{\rotatebox{90}{No-watermark}}} & \checkmark & \checkmark & \textbf{1.82}$\pm$0.006 & \textbf{2.44}$\pm$0.013 & \textbf{2.38}$\pm$0.013 & \textbf{2.32}$\pm$0.013 & \textbf{3.18}$\pm$0.028 & & &\\
     & \checkmark &  & \textbf{1.72}$\pm$0.007 & \textbf{2.31}$\pm$0.012 & \textbf{2.26}$\pm$0.012 & \textbf{2.17}$\pm$0.011 & \textbf{2.87}$\pm$0.025 & & &\\
     &  & \checkmark & \textbf{1.24}$\pm$0.003 & \textbf{1.17}$\pm$0.008 & \textbf{1.16}$\pm$0.009 & \textbf{1.19}$\pm$0.008 & \textbf{1.13}$\pm$0.009 & & &\\
     &  &  & \textbf{1.24}$\pm$0.003 & \textbf{1.20}$\pm$0.008 & \textbf{1.19}$\pm$0.008 & \textbf{1.21}$\pm$0.007 & \textbf{1.15}$\pm$0.009 & & &\\
     & $\Delta$ & \checkmark & \textbf{0.57}$\pm$0.006 & \textbf{1.27}$\pm$0.015 & \textbf{1.22}$\pm$0.015 & \textbf{1.13}$\pm$0.014 & \textbf{2.05}$\pm$0.028 & & &\\
     & $\Delta$ &  & \textbf{0.48}$\pm$0.006 & \textbf{1.11}$\pm$0.013 & \textbf{1.07}$\pm$0.014 & \textbf{0.97}$\pm$0.012 & \textbf{1.71}$\pm$0.026 & & &\\
\bottomrule
\end{tabular}
\end{small}
\caption{Means and standard errors (SE) of the relative importance on watermarks (RIW), quantifying the amount of importance attributed to pixels manipulated by insertion of a fixed-position translucent watermark in the watermark study, relative to the entire image. Data are shown for identical test images with and without the actual watermark applied (indicated in the WM column), for both test image classes (dogs and cats, Cat column), for different machine learning models (Model column), for which watermarks acted either as confounders (confounded setting) or non-informative features (balanced setting) during training, or were completely absent during training (no-watermark setting), and for feature attribution methods (Deconvolution, Integrated Gradients, Gradient SHAP, LRP-$\epsilon$, and LRP-$\alpha \beta$ columns) as well as model-independent baselines (Laplace and raw image $\mathbf{x}$ columns). Also shown are RIW differences between watermarked and corresponding non-watermarked images ($\Delta$-marked rows in WM column, where positive differences mark higher RIW on watermarked images).}
\label{tab:mean_se}
\end{table}

\begin{figure}[ht!]
    \centering
        \includegraphics[width=0.95\textwidth]{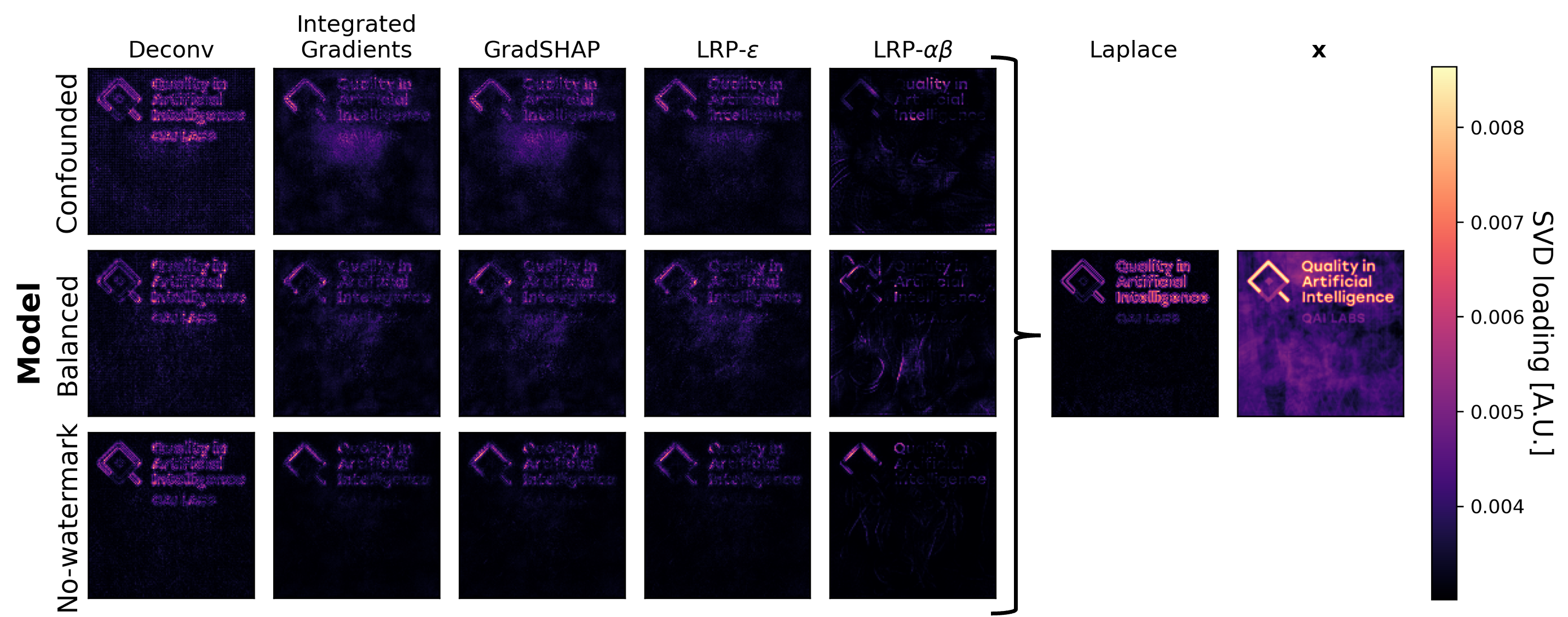}
    \caption{Absolute-valued first singular vectors of pixel $\times$ sample attribution matrices produced for cat and dog test images with watermarks in the watermark study using models trained on confounded, balanced, and no-watermark data (shown across rows) in combination with five different feature attribution methods (Deconvolution, Integrated Gradients, Gradient SHAP, LRP-$\epsilon$, and LRP-$\alpha \beta$) and two model-independent baselines ((Laplace and raw images $\mathbf{x}$, shown across columns). 
    The watermark is clearly visible for every combination of model training setting and attribution method, confirming the watermark as the most salient structure in all studied cases including cases where watermarks were present during training or represented non-informative nuisance features to the model. 
    Singular vectors obtained from attributions closely resemble singular vectors obtained from model-independent edge-detection filters (Laplace) and even the raw input image data ($\mathbf{x}$).}
    \label{fig:explanation-pca}
\end{figure}



\begin{table}[htbp]
\sffamily 
\centering
\begin{tabular}{lrrr}
\toprule
Model     &  \multicolumn{3}{c}{AUROC $\times$ 100 on Test Data} \\ 
 & \multicolumn{1}{c}{Confounded} & \multicolumn{1}{c}{Balanced} & \multicolumn{1}{c}{No-watermark} \\ 
 \midrule
Confounded & {\bftab 91.80} $\pm$ 0.008  & 79.32 $\pm$ 0.076 & 86.89 $\pm$ 0.011 \\
Balanced & 88.00 $\pm$ 0.016 & 87.75 $\pm$ 0.010 & {\bftab 88.76} $\pm$ 0.011 \\
No-watermark & 83.60 $\pm$ 0.031 & 85.00 $\pm$ 0.016 & {\bftab 89.72} $\pm$ 0.077 \\ 
\bottomrule
\end{tabular}
\caption{Area under the receiver operating characteristic curve (AUROC) dog vs. cat classification performance under different statistical image manipulations with variable-position watermarks. Machine learning models trained over confounded, balanced, and no-watermark images were applied to all three corresponding test sets, respectively. Results are averaged over five trained models for each of five random data splits into training, validation, and test sets, with standard deviations across models and splits shown.}
\label{tab:AUROCs-variable}
\end{table}

\begin{figure}[ht!]
    \centering
        \includegraphics[width=0.95\textwidth]{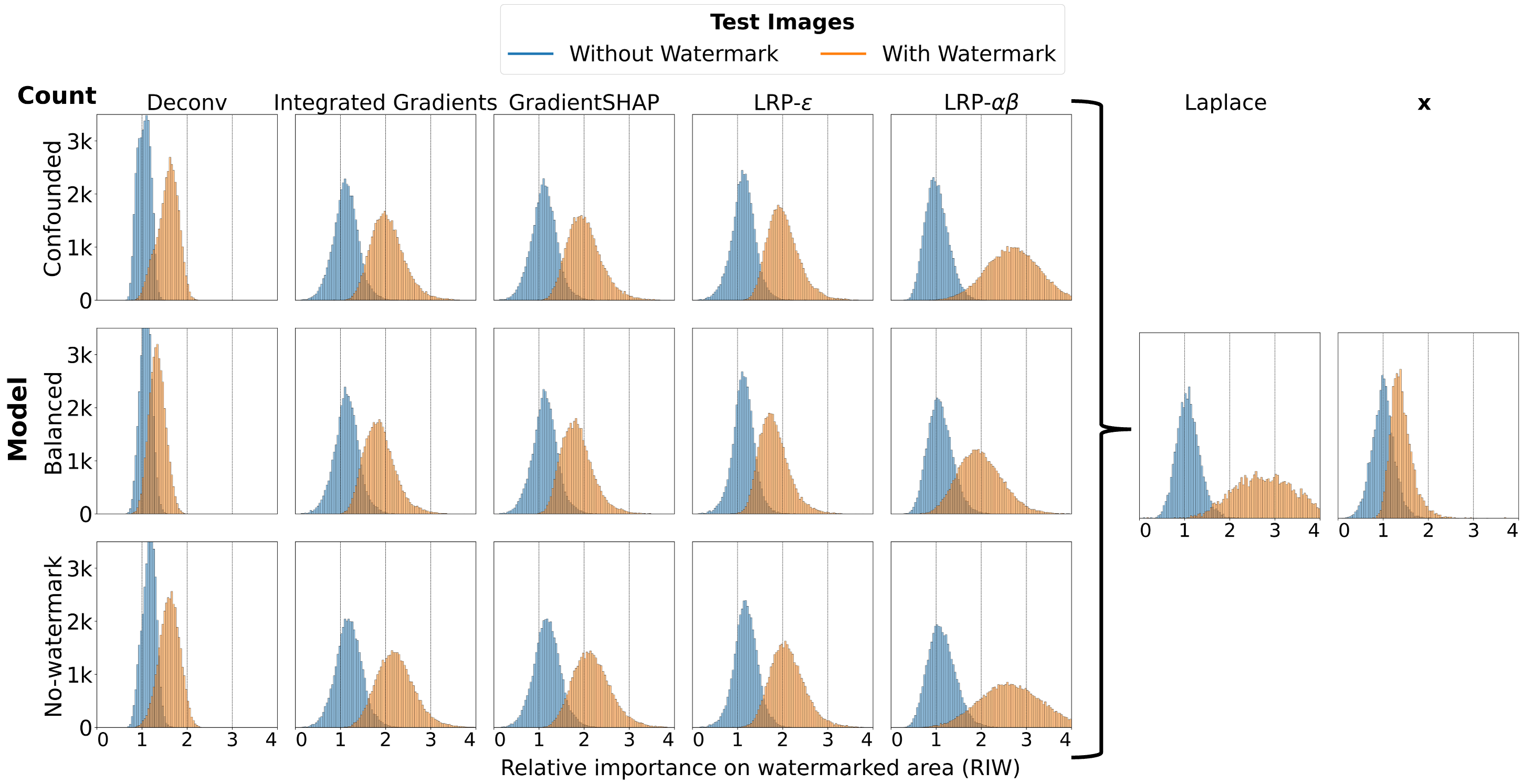}
    \caption{Distributions of the relative importance on watermarks (RIW), quantifying the amount of importance attributed to pixels manipulated by insertion of a variable-positioned translucent watermark in the watermark study, relative to the entire image. RIW distributions are shown for identical test images with and without the actual watermark applied, for different machine learning models (shown across rows) for which watermarks acted either as confounders (confounded setting) or non-informative features (balanced setting) during training or were completely absent during training (no-watermark setting), and for a selection of feature attribution methods (Deconvolution, Integrated Gradients, Gradient SHAP, LRP-$\epsilon$, and LRP-$\alpha \beta$) as well as two model-independent baselines (Laplace and raw images $\mathbf{x}$), shown across columns. 
    It can be seen that pixels manipulated by the watermark are attributed substantially higher RIW if a watermark is actually present compared to the case where no watermark is present. This is true across all studied attribution methods and baselines but also across all experimental settings including the balanced and no-watermark settings, in which the presence of a watermark is non-informative about the class label.
    Results are qualitatively and quantitatively similar to those obtained for fixed-position watermarks (c.f. Figure~\ref{fig:hist_watermark_fixed}).
    }
    \label{fig:hist_watermark_variable}
\end{figure}



\begin{table}[h!]
\sffamily
\centering
\begin{tabular}{lrrr}
\toprule
Model     &  \multicolumn{3}{c}{AUROC $\times$ 100 on Test Data} \\ 
 & \multicolumn{1}{c}{Confounded} & \multicolumn{1}{c}{Balanced} & \multicolumn{1}{c}{No-watermark} \\ 
 \midrule
Confounded & {\bftab 92.24} $\pm$ 0.007  & 78.15 $\pm$ 0.024 & 85.88 $\pm$ 0.013 \\
Balanced & 88.60 $\pm$ 0.017 & 88.43 $\pm$ 0.009 & {\bftab 89.03} $\pm$ 0.009 \\
No-watermark & 84.81 $\pm$ 0.034 & 86.84 $\pm$ 0.010 & {\bftab 89.54} $\pm$ 0.011 \\ 
\bottomrule
\end{tabular}
\caption{Area under the receiver operating characteristic curve (AUROC) dog vs. cat classification performance under different statistical image manipulations with fixed-position watermarks with inverted colour encoding. Machine learning models trained over confounded, balanced, and no-watermark images were applied to all three corresponding test sets, respectively. Results are averaged over five trained models for each of five random data splits into training, validation, and test sets, with standard deviations across models and splits shown.}
\label{tab:AUROCs-inverted}
\end{table}

\begin{figure}[ht!]
    \centering
        \includegraphics[width=0.95\textwidth]{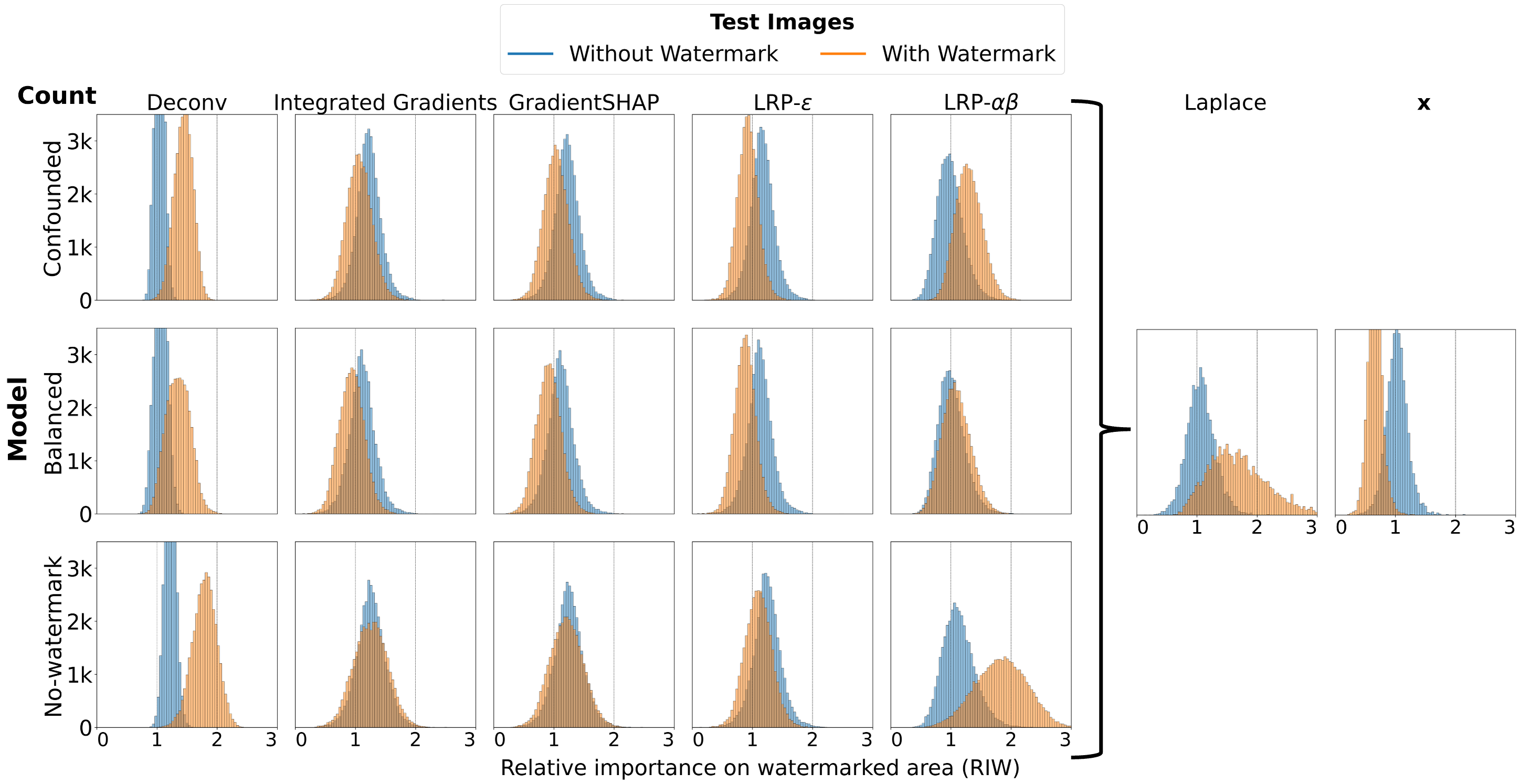}
    \caption{Distributions of the relative importance on watermarks (RIW), quantifying the amount of importance attributed to pixels manipulated by insertion of a fixed-position translucent watermark in the watermark study, relative to the entire image. This is done with all datasets using inverted colour encoding, where 1 encodes darkest and 0 encodes lightest regions. 
    RIW distributions are shown for identical test images with and without the actual watermark applied, for different machine learning models (shown across rows), for which watermarks acted either as confounders (confounded setting) or non-informative features (balanced setting) or were completely absent during training (no-watermark setting), and for a selection of feature attribution methods (Deconvolution, Integrated Gradients, Gradient SHAP, LRP-$\epsilon$, and LRP-$\alpha \beta$) as well as two model-independent baselines (Laplace and raw images $\mathbf{x}$), shown across columns.
    In contrast to results obtained using standard colour encoding, shown in Figures~\ref{fig:hist_watermark_fixed} and \ref{fig:hist_watermark_variable}, inverted colour encoding leads to substantially \emph{lower} RIW for watermarked than non-watermarked images for various attribution methods, even in the confounded setting, where the watermark carries genuine class-related information. An interesting case is given for LRP whose $\epsilon$ and $\alpha \beta$ variants display opposing behaviour in all studied settings.}
    \label{fig:hist_watermark_inverted}
\end{figure}

\end{document}